\documentclass[5p,times]{elsarticle}
\makeatletter
\def\ps@pprintTitle{%
 \let\@oddhead\@empty
 \let\@evenhead\@empty
 \let\@evenfoot\@oddfoot}
\makeatother

\usepackage[utf8]{inputenc}
\usepackage[binary-units]{siunitx}
\usepackage{mathtools}  
\usepackage{adjustbox}
\usepackage{amssymb}
\usepackage{amsmath}
\usepackage{natbib}
\usepackage{xcolor}
\usepackage{subcaption}
\usepackage{multirow}
\usepackage{graphicx}
\newsavebox{\imagebox}
\usepackage{multirow}
\usepackage{array}
\usepackage[hidelinks]{hyperref}
\usepackage{comment}
\usepackage{listings}
\usepackage[ruled,linesnumbered]{algorithm2e}

\begin{document}

\begin{frontmatter}

\title{
Federated learning framework for collaborative remaining useful life\\ prognostics: an aircraft engine case study
}

\author[1,3,4]{Diogo Landau}
\author[2,3]{Ingeborg de Pater}
\author[1]{Mihaela Mitici}
\author[1]{Nishant Saurabh}

\address[1]{Department of Information and Computing Sciences, Utrecht University,  Heidelberglaan 8, 3584 CS Utrecht}
\address[2]{Faculty of Aerospace Engineering, Delft University of Technology, Kluyverweg 1, HS 2926 Delft, The Netherlands}
\address[3]{These authors contributed equally to this work}
\address[4]{Corresponding author, d.hewittmouracarrapatolandau@uu.nl}

\begin{abstract}
Complex systems such as aircraft engines are continuously monitored by sensors. In predictive aircraft maintenance, the collected sensor measurements are used to estimate the health condition and the Remaining Useful Life (RUL) of such systems. However, a major challenge when developing prognostics is the limited number of run-to-failure data samples. This challenge could be overcome if multiple airlines would share their run-to-failure data samples such that sufficient learning can be achieved. Due to privacy concerns, however, airlines are reluctant to share their data in a centralized setting.  In this paper,  a collaborative federated learning framework is therefore developed instead. Here, several airlines  cooperate to train a collective RUL prognostic machine learning model, without the need to centrally share their data. For this, a decentralized validation procedure is proposed to validate the prognostics model without sharing any data. Moreover,  sensor data is often noisy and of low quality. This paper therefore proposes four novel  methods to aggregate the  parameters of the global prognostic model. These methods enhance the robustness of the FL framework against noisy data. The proposed framework is illustrated for training a collaborative RUL prognostic model for aircraft engines, using the N-CMAPSS dataset. Here, six airlines are considered, that collaborate in the FL framework to train a collective RUL prognostic model for their aircraft's engines. When comparing the proposed  FL framework with the case where each airline independently develops their own prognostic model, the results show that FL leads to more accurate RUL prognostics for five out of the six airlines. Moreover, the novel robust aggregation methods render the FL framework robust to noisy data samples.
\end{abstract}


\end{frontmatter}

\section{Introduction} \label{sec:intro}
The emergence of distributed computing technologies integrated with machine learning techniques, has gained traction for distributed data processing across various application domains. 
Predictive maintenance, and in particular the generation of data-driven Remaining Useful Life (RUL) prognostics for collaborative systems, 
has benefited in recent years from the integration of distributed computing technologies with machine learning models \cite{fink2020potential, mitici2023dynamic}.  

The main challenge when developing RUL prognostics using machine learning is the lack of run-to-failure data samples \cite{fink2020potential, de2023developing}. This is because, although systems are continuously monitored by sensors, they are often maintained/replaced preventively before their actual failure (preventive maintenance). This is, for example, the case for safety-critical aircraft systems such as the engines, where preventive maintenance is a means to limit the risk and consequences of unexpected failures. 
Most condition-monitoring data samples of such systems are unlabelled, i.e., the true RUL or health state is unknown. Only in a few cases are the systems monitored until their actual failure, i.e., there are very few run-to-failure data samples or, equivalently, there are very few labelled data samples for which the true RUL is known. 
This lack of run-to-failure data samples is a challenge when developing RUL prognostics using machine learning models. 

One way to overcome this challenge is for clients (airlines), each with their own 
condition-monitoring data samples, 
to collaborate in developing a common RUL prognostics model. When clients collaborate and share labeled data samples, valuable data are made available to train machine learning models for RUL prognostics. In the case of prognostics development for aircraft systems, however, the airlines are very hesitant to share their data, or to transmit the data for processing to a centralized server, due to privacy concerns and possible conflicts of interests \cite{zhang2021federatedknowdledge, AICHAIN, fed_atm}. %
The aversion to data sharing  remains even when the data is anonymized or processed \cite{zhang2021federatedknowdledge}. 

In this paper, {\color{black} a collaborative federated learning (FL) framework is proposed for developing RUL prognostics for aircraft systems.} For this purpose, a common machine learning RUL prognostic model is trained using the data of several airlines. It is assumed that each airline (client) possesses a local compute device. {\color{black}Collaborative FL is initially} deployed across these local devices  \cite{mcmahan2017communication}, {\color{black} similar to classic FL}, such that the clients (airlines) do not have to share or transmit data to a centralized server to train this common model. 
Firstly, it makes use of the intrinsic value of data samples of multiple airlines, further recognizing that the collective intelligence of multiple airline data samples can enable the development of accurate and robust RUL prognostics. Secondly, it aligns with the imperative to preserve data privacy, an issue of paramount importance for airlines. By taking condition monitoring data processing and prognostics model training to the individual airlines, this approach  minimizes data transfer, and empowers airlines to collaboratively generate RUL prognostics, while safeguarding sensitive operational information. 

From a methodological perspective, there are two main aspects to consider when developing a {\color{black}collaborative} FL framework for RUL prognostics for aircraft systems.
Firstly, when clients collaborate, the quality of the data has a direct impact on the final collaborative model.
For example, condition-monitoring data samples are often noisy due to interference from the outside environment, data transmission, various data sources and various sensors \cite{lei2018machinery, rezaeianjouybari2020deep}. Such noisy data directly impacts the accuracy of the common model \cite{bhagoji2019analyzing, zhang2021federatedknowdledge, guo2022fedrul, wang2020model}.  Standard aggregation procedures used for {\color{black}classic} FL, e.g., FedAvg \cite{mcmahan2017communication}, are unable to prevent clients with noisy data from participating in the development of a common model, leading to a decreased accuracy of the model. \textcolor{black}{To mitigate this, recent studies \cite{yin2018byzantine,blanchard2017machine} have provided robust algorithms that guarantee model convergence in the presence of noisy contributions. However, these methods assume that the data is independent and identically distributed (i.i.d.) between the different clients, which is rarely the case for airlines. For instance, each airlines flies different routes with their aircraft, which influences the sensor measurements. As such, this paper presents robust parameter aggregation procedures for {\color{black}a collaborative FL-based} RUL prognostics that are robust to noisy data samples.}

Secondly,  a decentralized validation procedure is formalized for validating a RUL prognostic machine learning model, embedded in the {\color{black} proposed} FL framework. This decentralized validation procedure is used to train the RUL prognostics models.
Generally, {\color{black} classic} FL is performed i) without any validation procedure \cite{zhang2021federatedmecha,  lu2023federated, chen2022federated, zhao2023federated, zhang2022data}, or ii) by using a global validation set \cite{chen2021federated, zhang2021federatedknowdledge, guo2022fedrul,fang2020local}. However, neither of these two cases is applicable in aviation.
First, the European Union Aviation Safety Agency (EASA) requires that the machine learning models are validated \cite{EASA} in order to approve their usage. Without a validation procedure, machine learning models for RUL prognostics of aircraft systems cannot be certified and implemented in practice. Second, the alternative of having a global validation set would require the airlines to share some data with the server, which violates their privacy. As such, assuming the availability of a global validation set for RUL prognostics of aircraft systems, as assumed in \cite{chen2021federated, zhang2021federatedknowdledge, guo2022fedrul}, is not realistic.  
To address this, the proposed decentralized validation procedure enables each client (airline) to validate the global model with their own validation dataset, and share only the loss obtained. With  this, a global validation loss is determined. Overall, the proposed decentralized validation procedure supports the certification of  machine learning models for RUL prognostics of aircraft systems, without requiring the airlines to share their data.


The main contributions of this paper are as follows:
\begin{itemize}
    \item A {\color{black}collaborative FL framework} is proposed to train a RUL prognostic machine learning model with multiple clients (airlines) without sharing any data.  For this,  a decentralized validation procedure is formalized to validate the machine learning models, which is embedded in the FL framework. As data privacy preservation is crucial in such settings, this decentralized validation procedure does not require the clients (airlines) to share any data or labels. Nevertheless, this decentralized validation procedure is generic, and can be employed for other applications of FL. 
    \item Four robust  approaches are developed to aggregate the parameters for the global model. This enables the development of accurate RUL prognostics even when some clients (airlines) have noisy data samples. 

    \item The proposed framework is applied to a  realistic case study, where 6 airlines collaborate to train a RUL prognostic model for aircraft engines. For this the N-CMAPSS dataset \cite{arias2021aircraft} on aircraft engines is used. 
        The proposed FL approach  is compared with the case when airlines do not collaborate at all and each airline independently trains its own RUL prognostic models. In this case, the proposed FL approach leads to more accurate RUL prognostics for 5 out of the 6 considered airlines. The complete source code of the proposed FL framework including the reproducibility and data artifact is released opensource in an \emph{online} \verb|GitHub| repository~\cite{rul-fl}.
        
   
\end{itemize}


The rest of the paper is organized as follows. 
In Section \ref{sec:fed_valid}, an overview of the FL approach is provided, which includes the decentralized validation procedure. In Section \ref{sec:method} this FL approach is further extended with noise-robust aggregation procedures for the model parameters. In Section \ref{sec:case_study} the proposed approach is illustrated in a case study where airlines collaboratively generate RUL prognostics. Numerical results are provided in Section \ref{sec:results}, and conclusions are provided in Section \ref{sec:Conclusion}.

\section{\textcolor{black}{Related Work}}
\label{sec:related_works}

\textcolor{black}{
Recent studies employed FL in a variety of applications, such as healthcare \cite{li2023review}, smart cities \cite{djenouri2023federated} and the Internet of Things \cite{al2023pelican, 10.1007/978-981-99-9836-4_12}. Most studies on FL in the domain of condition monitoring and health management of systems focus on fault diagnosis. For instance, in \cite{zhang2021federatedmecha, zhang2021federatedknowdledge, chen2022federated, zhao2023federated, yang2021federated, zhang2022data, lu2023federated}, the faults of bearings are diagnosed. }

\textcolor{black}{The development of a RUL prognostics model with FL is very rare in literature. In \cite{guo2022fedrul}, a convolutional autoencoder is trained at each client. The clients then send the encoded state of each data sample, together with the corresponding label to the server. This server trains a RUL prognostic model on these encoded states. For sensitive applications, however, the encoded state of the data samples and corresponding labels may already be too valuable to share. Therefore, in this paper, airlines simply share their local models that resulted from training on their local datasets. In \cite{lu2024federated}, FL is used to train a RUL prognostic model for clients with heterogeneous data. Here, the clients are first clustered based on the similarity of their data, and RUL prognostic models are trained based on these clusters. This paper, however, considers a problem where airlines train a common RUL prognostic model through FL. As not many airlines exist, and as each airlines only has very few labeled data samples, this RUL prognostic model is trained with all participating airlines.
}

\subsection{\textcolor{black}{Aggregation Methods}}

\textcolor{black}{
Given the increased statistical heterogeneity in FL compared to other distributed learning settings, aggregation methods have been a core concern in the literature. Li et al.~\cite{li2018federated} modify \textit{FedAvg} to account for statistical and system heterogeneity, while Ji et al.~\cite{ji2019learning} leverage a minimisation problem to compute the global model that is closest (in distance) to the submitted local models. However, these procedures do not account for noisy local models that negatively contribute to the global aggregation procedure, a scenario that if unhandled, deters potential participants.
}

\textcolor{black}{
For this reason, several robust aggregation procedures~\cite{alebouyeh2024benchmarking,hayes2018contamination,damaskinos2018asynchronous,xie2019zeno} have been proposed to mitigate the impact clients with noisy data can have on the performance of the global model. Given a set of models trained by local clients, robust aggregation methods aim to maintain the convergence of the global model regardless of the presence of noise~\cite{alebouyeh2024benchmarking,yin2018byzantine} or malicious intent~\cite{blanchard2017machine,xie2018generalized}. Blanchard et al.~\cite{blanchard2017machine} select a single model as the global model based on a distance metric computed for all local models. Yin et al.~\cite{yin2018byzantine} instead propose two methods that compute the global model based on the distribution of the parameters over all local models. Nevertheless, as mentioned by the authors and empirically evaluated in~\cite{fang2020local}, these methods assume the data is independent and identically distributed, an assumption that does not hold in the considered scenario where the data is held by different airline companies.
}

\textcolor{black}{
To circumvent this, Fang et al.~\cite{fang2020local} present an aggregation method that evaluates the performance of each local model on a \textit{global} validation dataset, followed by removing the models that have a negative impact on the global model's performance. Their use of a global validation dataset hinders its use in aviation, wherein multiple airlines would have to share data to construe a representative dataset for the server. For this reason, this paper presents a decentralized validation procedure to inform robust aggregation methods in the context of RUL prognostics.
}

\section{Federated Learning with a decentralized validation procedure} \label{sec:fed_valid} 

In this section, a FL approach with a decentralized validation procedure is proposed, where the clients do not have to share any data or data labels. 
\textcolor{black}{This approach is motivated by the validation procedure presented in \cite{li2021fedphp,yu2020salvaging,roder2022tracing,sannara2021federated}, where the clients possess a local test set to evaluate the performance of a given model. This approach is applicable to a broad set of machine learning problems (e.g. regression or classification problems).}

\textcolor{black}{Let $C$ be a set of clients (e.g., airlines, companies, transport companies), where each client $i \in C$ has its own dataset $X_i$. This dataset $X_i$ only belongs to client $i$, and cannot be shared with other clients or with the central server. In total, there are $|C| = N$ clients. } 
For instance, for RUL prognostics, the data set $X_i$ consists of data samples with sensor measurements, where each data sample has as label the actual RUL. 
Standard FL uses all data samples in the dataset $X_i$ of client $i \in C$ as training set. In contrast, in this paper the data samples at each client $i \in C$ are divided into a training set \textcolor{black}{$T_i \subset X_i$} and a validation set $V_i = X_i \setminus T_i$. \textcolor{black}{Thus, the training-validation split is performed at each client separately.}


\begin{figure}[!ht]
    \centering
        \includegraphics[width = 0.35\textwidth]{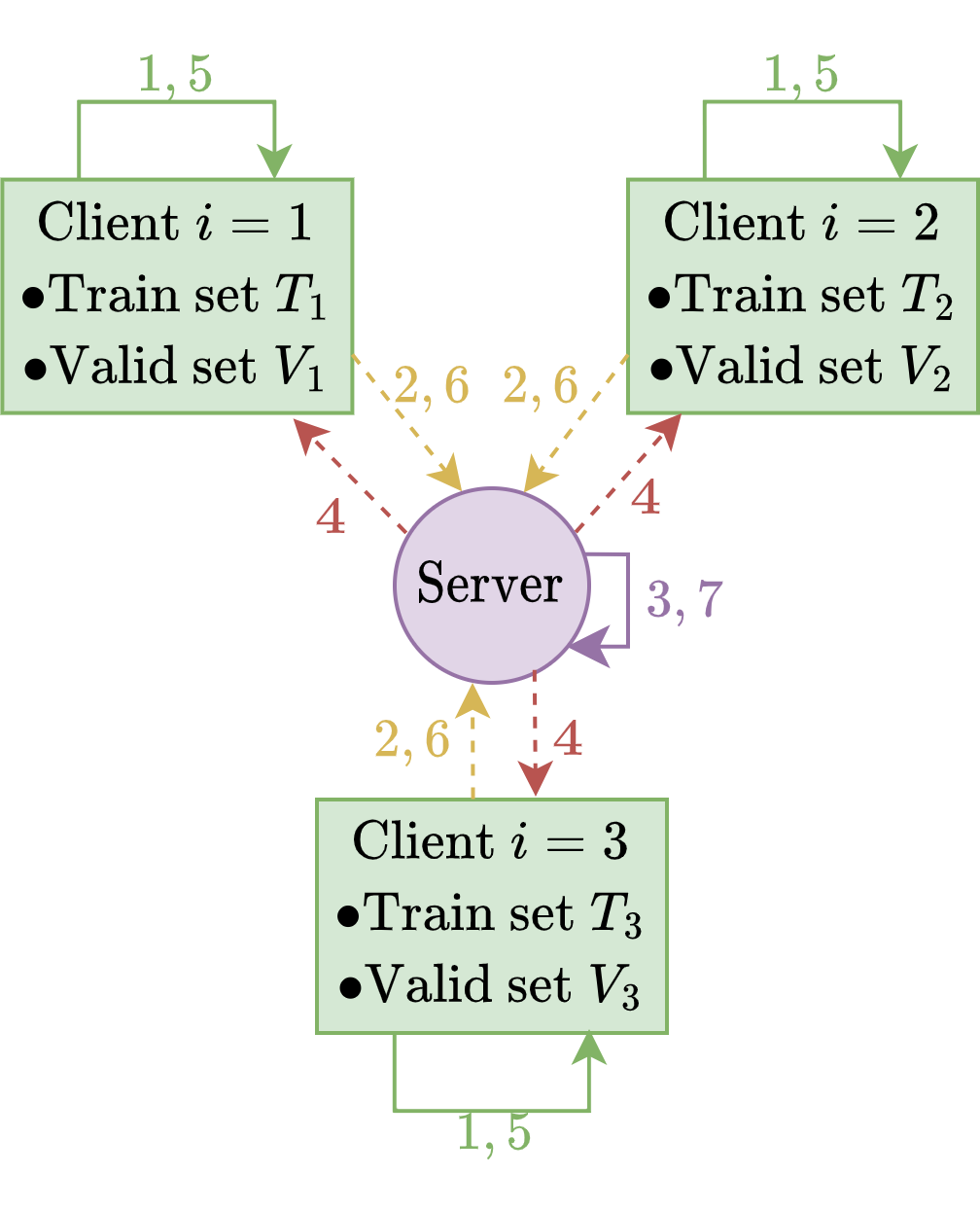}
          \caption{Schematic overview of FL with the proposed decentralized validation procedure, $N=3$ clients.}
        \label{fig:fl_procedure_all}
\end{figure}

\begin{figure}
    \centering
    \includegraphics[width =0.4 \textwidth]{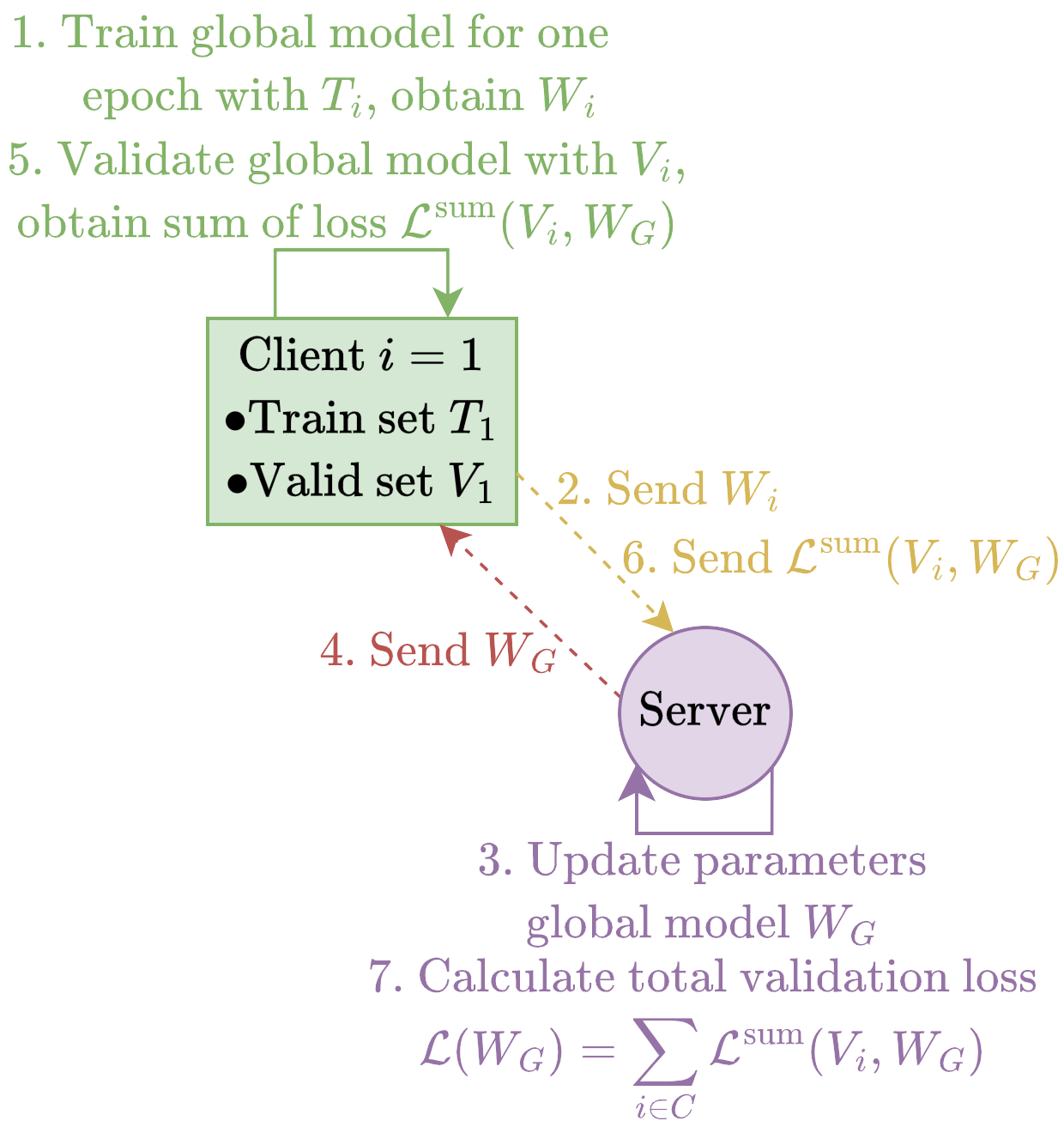}
    \caption{Zoom in - Schematic overview of all steps of the FL with the proposed decentralized validation procedure, zoomed in on one client and the server.}
    \label{fig:fl_procedure_one}
\end{figure}


The clients cooperate and communicate with each other through a central server, without sharing any data. This server contains the global model, which is trained jointly by all clients. In this Section, it is assumed that this global model is a neural network. Let  $W_G$ denote the parameters of this global model. It is also assumed that no global validation set is available at this server. 

The server first initializes the global model. The parameters $W_G$ are then sent to all clients $i \in C$. After this, the procedure for training the global model for one epoch with FL with the decentralized validation procedure is as follows (see also Figures \ref{fig:fl_procedure_all} and \ref{fig:fl_procedure_one}): 
\begin{enumerate}
    \item \label{item:train} Each client $i \in C$ trains the global model for one epoch using its training set $T_i$. The resulting neural network is referred to as the client's local model. Let $W_i$ denote the parameters of the local model of client $i \in C$. Note that only a part of the data ($T_i \subset X_i$) at each client $i \in C$ is used to train the model. 
    \item \label{item:wi_back} Each client $i \in C$ sends the parameters $W_i$ to the server. 
    \item \label{item:fedavg} The server aggregates the parameters $W_i$ from all clients $i \in C$, to generate the new global model. Each parameter $w_G \in W_G$ of the global model is a function $f(\cdot)$ of the corresponding parameter $w_i \in W_i$  of the local model of all clients $i \in C$, i.e.,: 
    \begin{equation} \label{eq:aggregation}
        w_G = f\left(w_1, w_2, \ldots, w_N\right). 
    \end{equation}    
    The standard aggregation procedure is called federated averaging (FedAvg). In  FedAvg, each parameter $w_G \in W_G$ of the global model is simply the weighted average of this parameter $w_i \in W_i$  over the models of all clients $i \in C$ \cite{mcmahan2017communication}:
    \begin{equation}
    \label{eq:fedavg}
        w_G =  f\left(w_1, w_2, \ldots, w_N\right) = \sum_{i \in C} \frac{|T_i|}{n} w_i,
    \end{equation}    
    where $|T_i|$ is the number of samples in the training set $T_i$ of client $i \in C$, while $n$ is the total number of samples at all clients, i.e., $n = \sum_{i \in C} |T_i|$. 
    \item The server sends the updated parameters $W_G$ of the global model to all clients $i \in C$. 
     \item Each client $i \in C$ validates the global model (with parameters $W_G$) with its own validation dataset $V_i$, by computing the sum of the loss $\mathcal{L}^{\text{sum}}(V_i, W_G)$. \textcolor{black}{This is, for instance, the sum of the squared errors for regression problems, or the sum of the cross entropy loss for classification problems. }
     \item Each client $i \in C$ sends the loss $\mathcal{L}^{\text{sum}}(V_i, W_G)$ back to the server. 
    \item \label{item:validate} The performance of the global model is assessed by calculating the total validation loss $\mathcal{L}(W_G)$ from the losses $\mathcal{L}^{\text{sum}}(V_i, W_G)$ of all clients $i \in C$:
      \begin{equation} \label{eq:global_validation_loss}
          \mathcal{L}(W_G) = \sum_{i \in C} \mathcal{L}^{\text{sum}}(V_i, W_G).
      \end{equation} 
\end{enumerate}
Note that Steps 5, 6 and 7 are not present in a {\color{black}{classical}} FL procedure without any validation. 

The model is trained until a certain stopping criterion is fulfilled, for example based on the number of epochs or on the total validation loss. Due to the decentralized validation procedure, after training, the parameters of the global model that result in the lowest validation loss can be selected.


\section{Noise-robust Federated Learning aggregation methods for model parameters}
\label{sec:method}
 
In Section \ref{sec:fed_valid}, Steps \ref{item:train} - \ref{item:validate} illustrate how a global model is generated and validated in a general distributed FL setup. With classical aggregation methods such as FedAvg, Step \ref{item:fedavg} computes a global model based on a weighted average of the local models provided by each client (Equation \ref{eq:fedavg}). With this approach, the contribution each model has in the global model is proportional to the size of the dataset used by each client. However, given this collaborative FL setup, it is possible that clients train their models on noisy data due to, for example, faulty sensors. In these conditions, due to the averaging performed by FedAvg, the noise has an adverse effect on the global model's performance. To mitigate such issues, in this Section, Step \ref{item:fedavg} is split into Steps \ref{item:fedavg}a and \ref{item:fedavg}b to add an evaluation procedure before computing the global model. The intuition behind this design, is that by evaluating the performance of the models before aggregation, a model's contribution to the global model can be tuned based on its performance on different datasets available at each client. Specifically, the weights associated with each model are tuned to increase the contribution of the best performing models, and decrease the contribution of the models that do not perform as well.

With this in mind, four parameter aggregation methods are presented to increase the resilience of the global model against noisy data. These aggregation methods are based on the proposed decentralized validation procedure. Here, the local validation sets are used to evaluate the importance of each client's local model for the aggregation procedure. The evaluation is performed every time a new global model is to be created. For this, step \ref{item:fedavg} in the FL framework (Section \ref{sec:fed_valid}) is split in two sub-steps: 
\begin{itemize}
    \item Step \ref{item:fedavg}a: First, it is evaluated how well a local model of client $i \in C$ (with parameters $W_i$) performs, based on the local validation sets. 
    \item Step \ref{item:fedavg}b: A new function $f(w_1, w_2, \ldots$ $, w_N)$ is considered to compute a parameter $w_G \in W_G$ of the global model from the corresponding parameters $w_i \in W_i$ in the local models of all clients $i \in C$. This function is based on how well the model of each client performs in Step \ref{item:fedavg}a.
\end{itemize}
\textcolor{black}{Data heterogeneity is commonly observed in classic FL setups. This means that, often, clients with larger datasets have a bigger footprint on the global model. However, instead of associating each model with an importance proportional to the size of their datasets, the proposed aggregation procedure computes a local model's contribution to the global model based on its performance on the local validation dataset(s).}

\subsection*{Step \ref{item:fedavg}a: Evaluating the performance of the local models} \label{sec:allocation}
In Step \ref{item:wi_back}, all clients send their local model to the federated learning server after training on their local datasets. In Step \ref{item:fedavg}a, the server is now interested in evaluating each local model's performance on another client's local validation dataset. As such, the server coordinates the clients that shall evaluate each local model, and distributes these models to these clients. Since the server is only interested in the model's performance on another client's dataset, each client provides the loss their assigned model achieved against its validation dataset. The fact that the server is coordinating the distribution of the local models also limits the possibility of identifying the client that created the model under evaluation.


\paragraph{i) Full validation policy}

\begin{figure}[!ht]
    \centering
     \includegraphics[width = 0.49\textwidth]{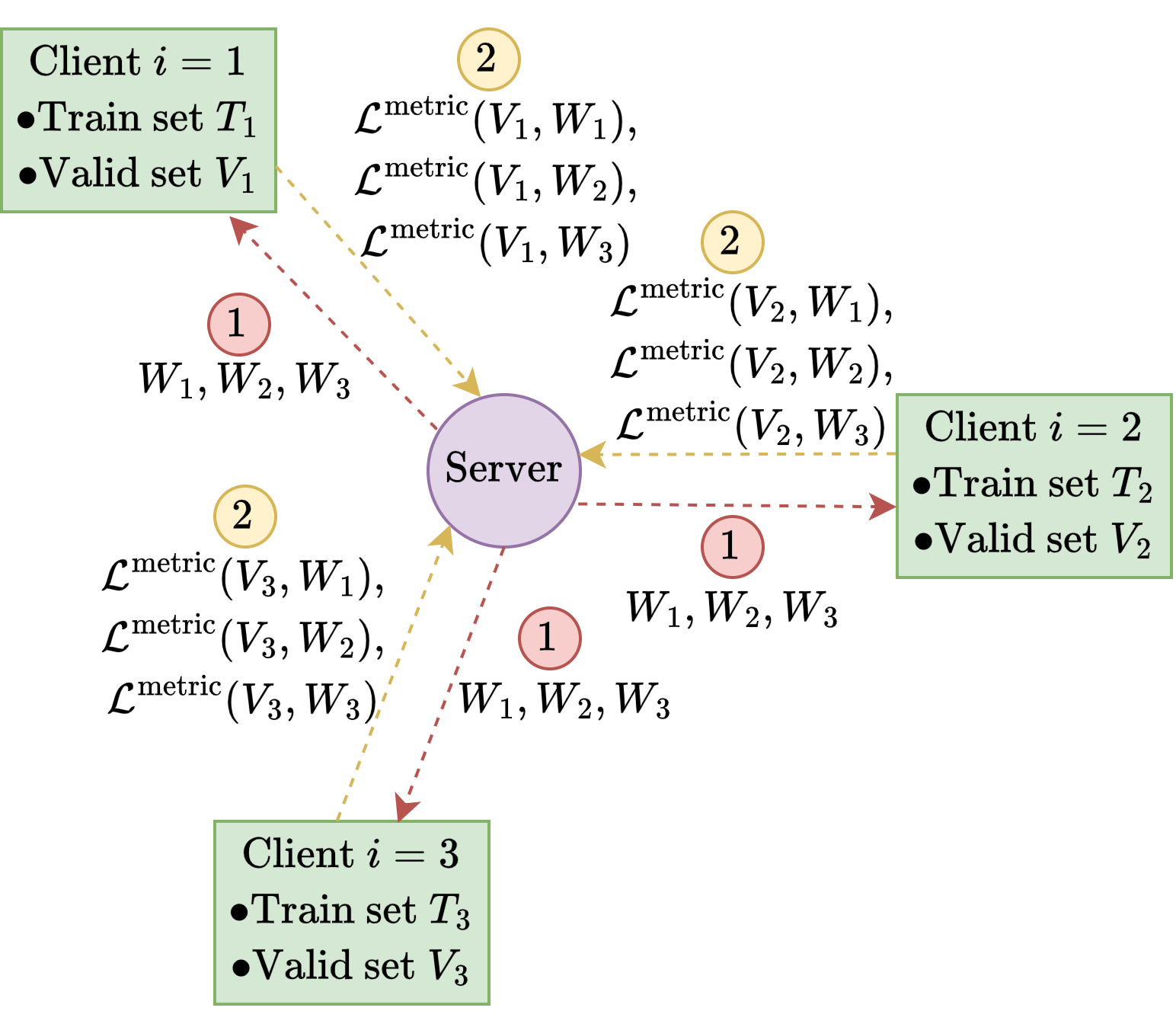}  
    \caption{Schematic illustration of the full validation policy, where (1) denotes the first step and (2) the second step in the communication sequence.}
    \label{fig:full_allocation_policy}
\end{figure}

\begin{figure}[!ht]
    \centering
        \begin{subfigure}{0.23\textwidth}
        \includegraphics[width = \textwidth]{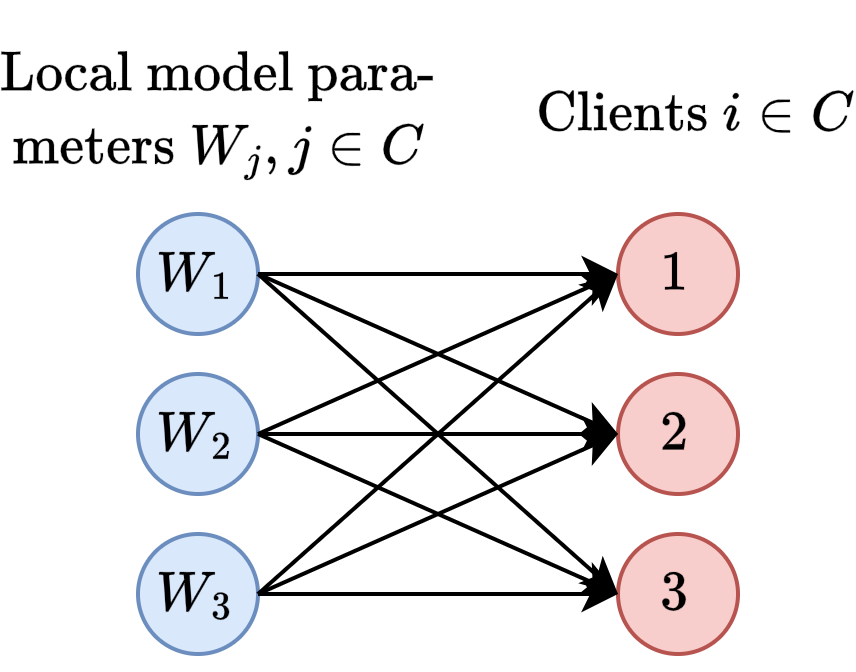}
    \caption{The full allocation. }    \label{fig:full_allocation}
    \end{subfigure}
    \begin{subfigure}{0.23\textwidth}
        \includegraphics[width = \textwidth]{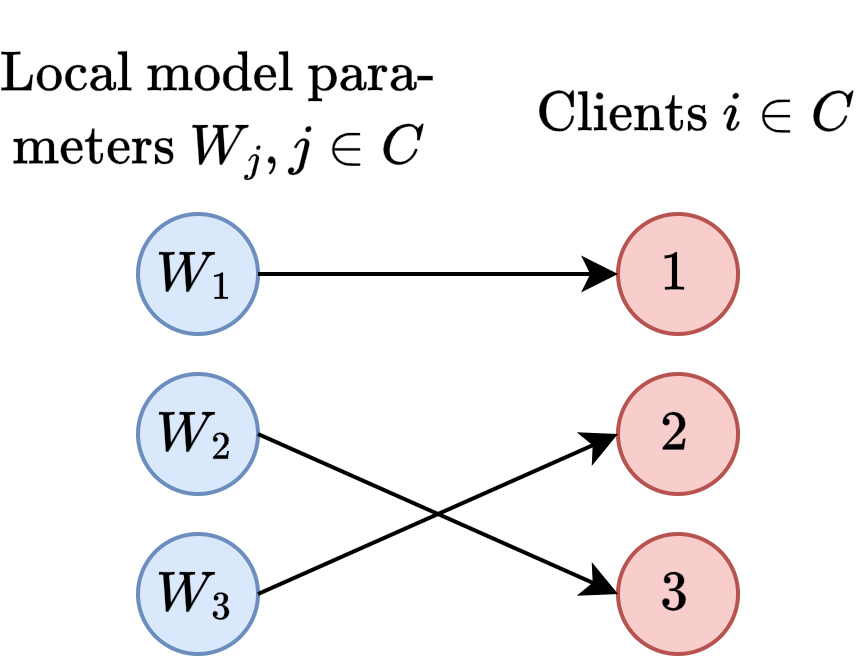}
    \caption{The random allocation. }    \label{fig:random_allocation}
    \end{subfigure}
    \caption{\textcolor{black}{An example of a full and random allocation with three clients.} }  
\end{figure}

Under the full validation policy, the server sends for each client $j \in C$, the parameters $W_j$ of the local model of client $j$ to all clients $i \in C$ (see Figure \ref{fig:full_allocation}). \textcolor{black}{Each client $i \in C$ then computes the loss $\mathcal{L}^{\text{metric}}(V_i, W_j)$ of the local model of client $j$, with parameters $W_j$, on its own validation set $V_i$ using a standard metric. This metric could be, for instance, the (Root) Mean Squared Error (for regression problems), or the Cross Entropy loss (for classification problems).} Each client $i \in C$ then sends this loss $\mathcal{L}^{\text{metric}}\left(V_i, W_j\right)$ of the local model of client $j$ back to the server. This full validation policy is illustrated in Figure \ref{fig:full_allocation_policy}.

The server then computes the evaluation score $E_j$ of client $j$, that represents the accuracy of the local model of client $j$. This evaluation score is the median of all  losses $\mathcal{L}^{\text{metric}}(V_i, W_j)$ over all clients $i \in C$:
\begin{equation}
\label{eq:model_evaluation_score_full}
    E_j = \text{median}_{i \in C} \left\{ \mathcal{L}^{\text{metric}}\left(V_i, W_j\right)\right\}.
\end{equation}
Taking the median of all losses, rather than the average or sum, decreases the influence of clients with noisy data on the evaluation score. A low evaluation score $E_j$ for a client $j$ implies that the model of client $j$ performs well.  The pseudocode for the full validation policy is in Algorithm \ref{alg:full}.   

\begin{algorithm}
    \caption{Step \ref{item:fedavg}a - The full validation policy} \label{alg:full}
     \For{\textup{Client} $j \in C$}{    
        \For{\textup{Client} $i \in C$}{
            Send the  parameters $W_j$ to client $i$\;
            Calculate the loss $\mathcal{L}^{\text{metric}}(V_i, W_j)$ with i) the parameters $W_j$ of the local model of client $j$ and ii) the validation set $V_i$ of client $i$\;
            Send the loss $\mathcal{L}^{\text{metric}}(V_i, W_j)$ back to the server\;
        }
        Calculate the evaluation score:  $E_j = \text{median}_{i \in C} \left\{ \mathcal{L}^{\text{metric}}\left(V_i, W_j\right)\right\}$\;
    }   
\end{algorithm}

\paragraph{ii) Random validation policy}

\begin{figure}[!ht]
    \centering
     \includegraphics[width = 0.49\textwidth]{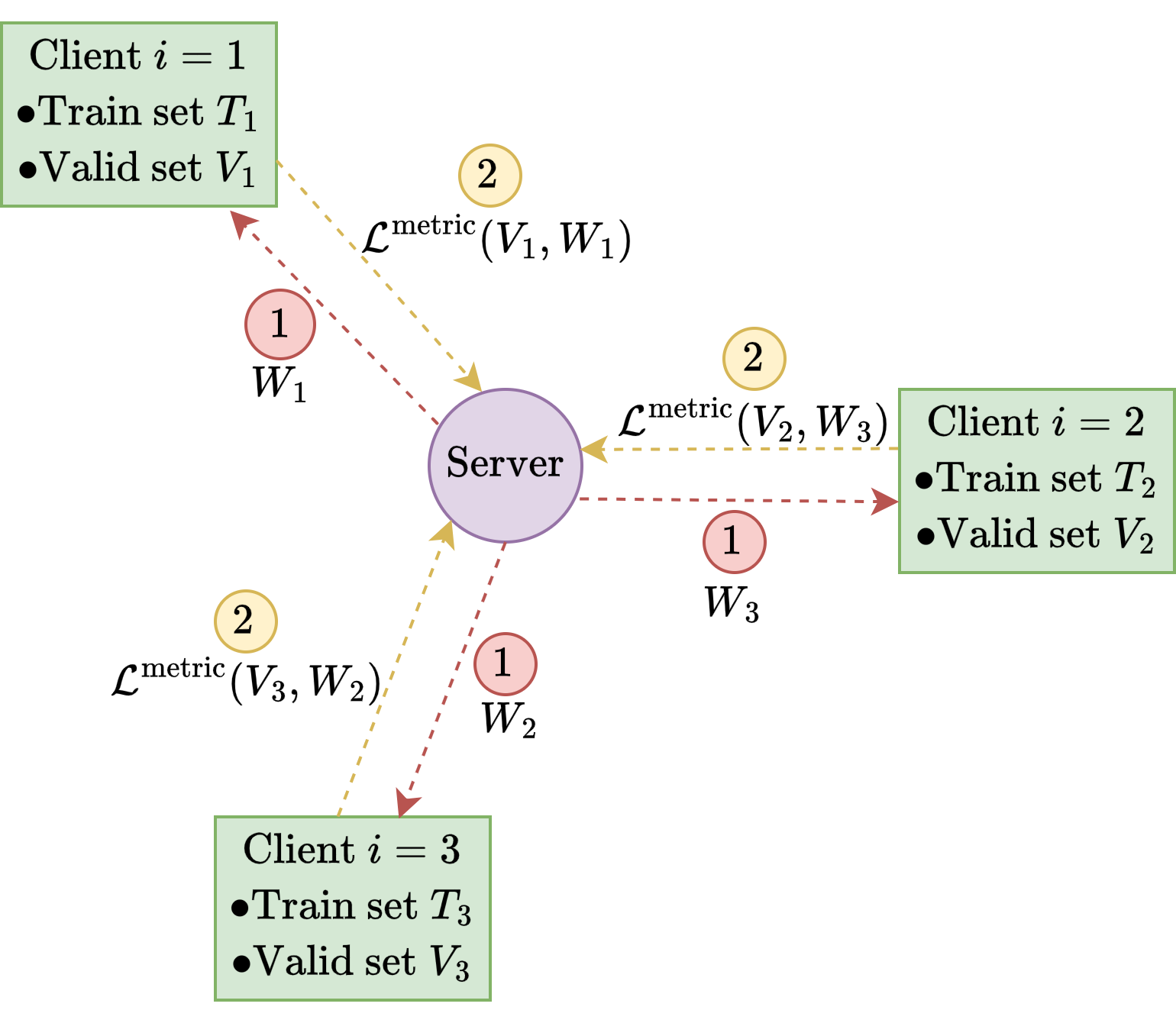}  
    \caption{Schematic illustration of the random validation policy. In this example, the weights of the clients are allocated as in Figure \ref{fig:random_allocation}.(1) denotes the first step and (2) the second step in the communication sequence.}
    \label{fig:random_allocation_policy}
\end{figure}

The full validation policy is relatively time-consuming, since the validation loss of all local models is computed with the local validation set of all clients.  The random validation policy therefore computes the  loss of each local model on only one validation set $V_i$ at a client $i \in C$. This random validation policy is illustrated in Figure \ref{fig:random_allocation_policy}.

The pseudocode for the random validation policy is in Algorithm \ref{alg:random}. The server first randomly allocates one local model of a client $j$, with parameters $W_j$, to each client $i \in C$. Each local model (with parameters $W_j$) is thus allocated to exactly one client $i \in C$, and reversely each client $i \in C$ has one local model (with parameters $W_j$) allocated to it. Let $i^{\text{ass}}_j \in C$ denote the client to which the local model of client $j$ is assigned. An illustration of such an allocation with three clients is depicted in Figure \ref{fig:random_allocation}.

For each client $j \in C$ the loss $\mathcal{L}^{\text{metric}}(V_{i^{\text{ass}}_j}, W_j)$ of the local model (with parameters $W_j$) is computed on the validation set  $V_{i^{\text{ass}}_j}$ of the assigned client $i^{\text{ass}}_j \in C$.  The evaluation score of client $j$ is then simply this loss:
\begin{equation} \label{eq:model_evaluation_score_random}
    E_j = \mathcal{L}^{\text{metric}}(V_{i^{\text{ass}}_j}, W_j).
\end{equation}

\begin{algorithm}[!ht]
    \caption{Step \ref{item:fedavg}a -  The random validation policy} \label{alg:random}
     Randomly assign the local model of each client $j \in C$ to one client $i \in C$. Ensure that exactly  one local model is assigned to each client $i \in C$.  Let $i^{\text{ass}}_j \in C$ denote the client to which the local model of client $j$ is assigned\;
     \For{\textup{Client} $j \in C$}{    
        Send the  parameters $W_j$ to the assigned client $i^{\text{ass}}_j$\;
         Compute the loss $\mathcal{L}^{\text{metric}}(V_{i^{\text{ass}}_j}, W_j)$ with i) the parameters $W_j$ of the local model of client $j$ and ii) the validation set $V_{i^{\text{ass}}_j}$ of client $i^{\text{ass}}_j$\;
         Send the loss $\mathcal{L}^{\text{metric}}(V_{i^{\text{ass}}_j}, W_j)$ back to the server\;        
        Set the evaluation score:  $E_j = \mathcal{L}^{\text{metric}}(V_{i^{\text{ass}}_j}, W_j)$\;
    }   
\end{algorithm}

\subsection*{Step \ref{item:fedavg}b: Aggregating the model parameters} \label{sec:aggregation} 

In step \ref{item:fedavg}b, the parameters of the global model $W_G$ are computed, based on the parameters $W_i$ and the evaluation scores $E_i$ of the local models of the clients $i \in C$. Here, two methods to aggregate the parameters are proposed: i) The best model aggregation policy, which selects the best model of all local models as the new global model and ii) the softmax aggregation policy, which uses a weighted average to compute the parameters of the global model.


\paragraph{Best model aggregation policy} 
The best model aggregation policy is an adaptation of the work of Blanchard et al. \cite{blanchard2017machine} that selects the best performing model based on an evaluation heuristic. The adaptation considered in this paper consists of modifying the evaluation procedure to use the random and full validation policies presented in Step \ref{item:fedavg}a. As a result, based on the evaluation score provided by the validation policy, the model that has the best evaluation score is selected as the new global model, as follows:
\begin{align}
    i^{\text{best}}  &= \text{arg min}_{i \in C} \{E_i\}, \\ 
    W_G &= W_{i^{\text{best}}}, \label{eq:aggregation_best_model}
\end{align}
where $i^{\text{best}}$ is the client with the lowest evaluation score.

\begin{figure}
    \centering
    \includegraphics[width=0.49\linewidth]{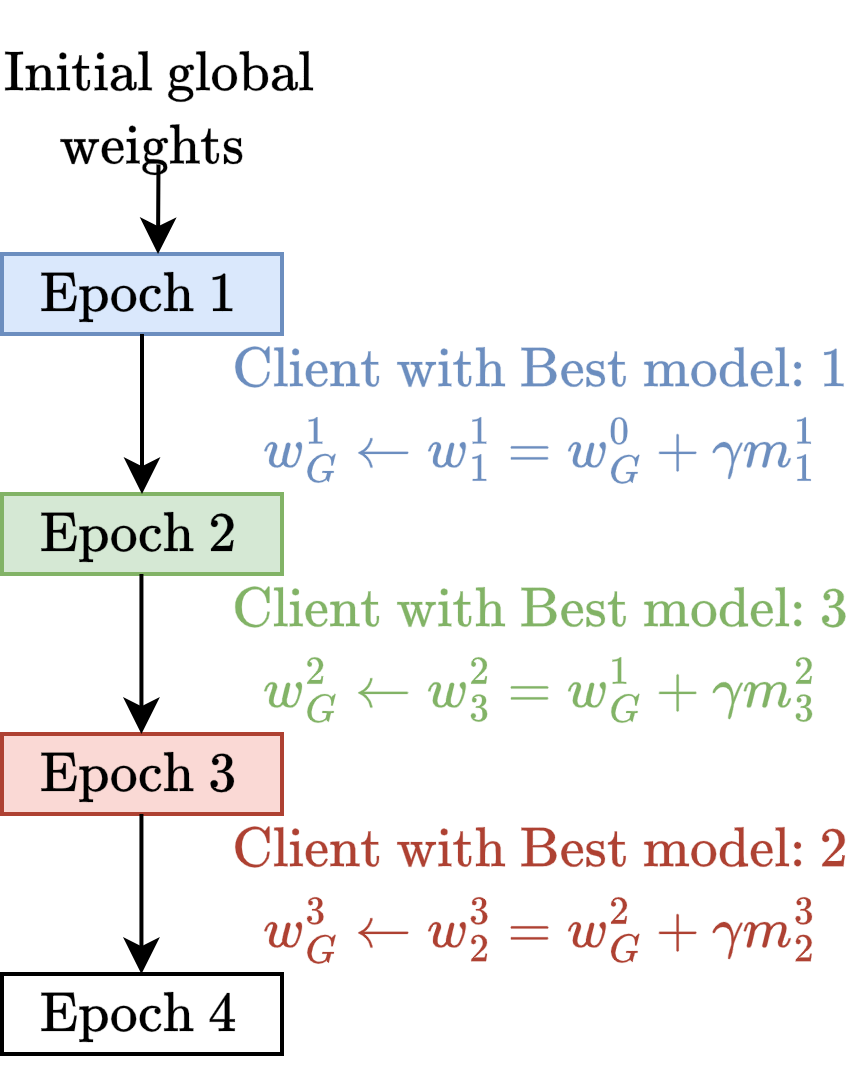}
    \caption{\textcolor{black}{Example of the \emph{Best model} aggregation policy with the  gradient descent algorithm. Here, $\gamma$ denotes the learning rate, $w^e_i$ denotes  a single model parameter at epoch $e$ and client $i$, $w_G^e$ denotes the corresponding single parameter weight at epoch $e$ for the global model,  and $m^e_i$ denotes the gradient of this parameter at epoch $e$ and client $i$.  }}
    \label{fig:best_agg}
\end{figure}

\textcolor{black}{After each epoch, the weights of only a single model are  selected as weights of the global model. However, the client with the best model differs from epoch to epoch. All clients, that are the client with the best model at least once, thus contribute to the final global weights through the weight updating algorithm. This is illustrated in Figure \ref{fig:best_agg} for the standard gradient descent algorithm. Here, $\gamma$ denotes the learning rate, and $m^e_i$ denotes the gradient of epoch $e$ at client $i$. After epoch 1, client 1 is selected as  client with the best model, and the global weights are updated with the gradients of client 1. In epoch 2,  however, client 3 is selected as client with the best model, and the gradients of client 3 are used to further update the weights. At epoch 3, the gradients of client 2 are used to update the weights. In this way, all three clients contribute to the final weights of the model.}

\paragraph{Softmax aggregation policy}
For the softmax policy,  the parameters of the global model are computed as a weighted average of the parameters of the local model. In this average, each client $i \in C$ gets a weight $\alpha_i$, based on the evaluation score $E_i$. 

A local model performs  better if the evaluation score is lower. First, the inverse of the evaluation score $A_i$ for each client $i \in C$ is therefore computed: 
\begin{alignat}{3}
    &A_i = \frac{1}{E_i} \label{eq:inverse_evaluation}
\end{alignat}
Then, the evaluation scores are normalized to diversify the final weights, after the softmax function. First, the mean $\mu$ and the standard deviation $\sigma$ of the inverse evaluation scores are computed:
\begin{alignat}{3}
    &\mu = \frac{1}{N} \sum_{i \in C} A_i \label{eq:mu}\\ 
    &\sigma = \sqrt{\frac{1}{N-1} \sum_{i \in C} \left(A_i - \mu\right)^2} 
\end{alignat}
With this,  the z-score $Z_i$ for each client $i \in C$ is computed:
\begin{alignat}{3}
     &Z_i = \frac{A_i - \mu}{\sigma} 
    \label{eq:normalize_zscore}  
\end{alignat}
Last, the weight $\alpha_i$ of each client $i \in C$ with the softmax function is computed:
\begin{alignat}{3}
     &\alpha_i = \frac{
        \mathrm{exp}(Z_i)
    }{
        \sum_{j \in C} \mathrm{exp}(Z_j)
    }.
    \label{eq:softmax}
\end{alignat}
The softmax function ensures that each weight is between 0 and 1, and that the sum of the weights is 1 (i.e., $\sum_{i \in C} \alpha_i = 1$). Ultimately,  a parameter $w_G \in W_G$ of the global model is computed with by taking the weighted average of the corresponding parameter $w_i \in W_i$ of the local models of the clients $i \in C$: 
\begin{alignat}{3}
    &w_G = \sum_{i \in C} \alpha_i w_i.
    \label{eq:softmax_global}
\end{alignat}

The Z-score is a common normalisation technique \cite{fei2021z}, and is used in this paper to obtain weights that better reflect the difference in evaluation scores ($E_i$). As an example, if airline $1$'s model has the validation score $E_1 = 20$, and thus $A_1 = \frac{1}{20} = 0.05$, and airline $2$'s model has the validation score $E_2 = 25$ and thus $A_2 = \frac{1}{25} = 0.04$, then the weight attributed to each model using $A_1$ and $A_2$ directly as input to the softmax function, is respectively $0.5025$ and $0.4975$. In this case both models  have an approximately equal contribution to the global model, despite the significant difference in their evaluation scores. To avoid this, it is important to normalise the range obtained by the evaluation score using Equation \ref{eq:normalize_zscore} corresponding to the Zscore computation. As a result, after computing the mean and standard deviation, $Z_1 = 0.707$ and $Z_2 = -0.707$, and using these values as input to the softmax function, the airline models resulting weights are $\alpha_1 = 0.804$ and $\alpha_2 = 0.196$. As such, using the Z-score normalisation has allowed a better differentiation of the weights attributed to each airline's models, which is proportional to their relative performance.



\subsection*{Novel aggregation methods for the global model parameters} \label{sec:agg_overview}
The two validation policies (Full validation and Random validation) and two aggregation policies (Best Model and Softmax) are proposed. The combination of these policies give four novel aggregation methods, as specified in Table \ref{tab:aggregation_algorithms}. These aggregation methods all replace step \ref{item:fedavg} of the FL framework in Section \ref{sec:fed_valid}. 

\begin{table}[!ht]
\caption{
    The four proposed aggregation methods based on the validation policy and the aggregation policy.  
}
\label{tab:aggregation_algorithms}
\centering
    \begin{tabular}{l|ll}
        \multicolumn{1}{c|}{\textbf{Name}} & \multicolumn{1}{c}{\textbf{Validation}} & \multicolumn{1}{c}{\textbf{Aggregation}} \\ 
         \multicolumn{1}{c|}{\textbf{method}} & \multicolumn{1}{c}{\textbf{policy}} & \multicolumn{1}{c}{\textbf{policy}} \\ \hline
        Random-Best                              & Random                                          & Best model \\
        Random-Softmax                           & Random                                         & Softmax\\
        Full-Best                                & Full                                           & Best model\\
        Full-Softmax                             & Full                                           & Softmax\\
     
    \end{tabular}
\end{table}




\section{Case Study: A multi-airline Federated Learning framework for RUL prognostics}
\label{sec:case_study}

The proposed FL framework with a decentralized validation procedure is illustrated for the case of  airlines collaborating to generate a common machine learning model for RUL prognostics of aircraft engines.
{\color{black}
An airline operates several aircraft. The engines of these aircraft are continuously monitored due to their safety criticality. The obtained measurements are used to learn degradation patterns and anticipate engine failures. To enhance this learning, airlines are interested in learning from other airlines' datasets, provided that the data privacy is preserved.}

The {\color{black}proposed FL framework is illustrated} using the condition monitoring data \textcolor{black}{samples} of aircraft engines in folder DS02 of the N-CMAPSS dataset \cite{arias2021aircraft}. 
The N-CMAPSS dataset consists of both 
sensor measurements and operating conditions of several engines. These measurements are available from the first flight performed with the engine, until the failure of the engine. Dataset DS02 consists of six engines (engines 2, 5, 10, 16, 18 and 20) whose data is meant for training, and three engines (engines 11, 14 and 15) to be used for testing.

{\color{black}
In this study, a case is considered where each airline has the data of one engine. Specifically, the training engines are  available at Airlines A-F as follows:  Airline A (engine 2), Airline B (engine 5), Airline C (engine 10), Airline D (engine 16), Airline E (engine 18), Airline F (engine 20).
The data of these six training engines is used to train a RUL prognostic model. 
The  test engines 11, 14, 15 are not assigned to a specific airline. They are instead used to analyze the performance of the RUL prognostic model by predicting the RUL of the test engines 11, 14 and 15 after each flight.
With this,  the performance of the proposed FL framework and the robust aggregation methods are analyzed.
}


Table \ref{tab:info_engines} gives a detailed overview of the \textcolor{black}{considered airlines and their engines}. The flights of each engine belong to one flight class: flight class 1 (short flights), flight class 2 (medium long flights) or flight class 3 (long flights). 

\subsection{Data preprocessing} \label{sec:data}

\begin{table}[!ht]
    \centering
        \caption{ Selected sensors and operating conditions which are used as input to the RUL prognostic model, 
        LPC - Low Pressure Combustor. HPC - High Pressure Combustor. HPT - High Pressure Turbine. LPT - Low Pressure Turbine. }
    \begin{tabular}{c|cc}
         Symbol & Description & Unit  \\
         \hline
         \multicolumn{3}{c}{Selected sensors} \\ 
         \hline 
         Wf & Fuel flow & pps \\ 
          Nf & Physical fan speed & rpm \\
          T24 & Total temperature at LPC outlet & $^{\circ}$R\\
          T30 & Total temperature at HPC outlet & $^{\circ}$R\\
          T48 & Total temperature at HPT outlet &  $^{\circ}$R\\ 
          T50 & Total temperature at LPT outlet &  $^{\circ}$R\\  
          P2 & Total pressure at fan inlet & psia \\ 
          P50 & Total pressure at LPT outlet & psia \\ 
          W21 & Fan flow & pps \\
          W50 & Flow out of LPT & lbm/s \\
          SmFan & Fan stall margin & - \\ 
          SmLPC & LPC stall margin & - \\ 
          SmHPC & HPC stall margin & - \\
          \hline
         \multicolumn{3}{c}{Operating conditions} \\ 
         \hline
         Mach & Flight Mach number & -\\ 
         TRA & Throttle-resolver angle & \%\\ 
         alt & Altitude of the aircraft & ft\\ 
         T2 & Total temperature at the fan inlet & $^{\circ}$R\\ 
         \hline 
    \end{tabular}
    \label{tab:selected}
\end{table}

\begin{table*}[!ht]
    \centering
        \caption{An overview of the airlines and their engines. Airlines A-F each have one engine for which measurements are recorded until the actual failure of each engine.  Fault mode 1: Degradation in the efficiency of the High Pressure Turbine (HPT). Fault mode 2: Degradation in the efficiency of the HPT, and in the efficiency and flow of the Low Pressure Turbine (LPT). }
    \begin{tabular}{c|cccccc|ccc}
        & \multicolumn{6}{c|}{Training} & \multicolumn{3}{c}{Testing} \\
        \hline 
          \textcolor{black}{Airline (engine)} & \textcolor{black}{A (2)} & \textcolor{black}{B (5)} & \textcolor{black}{C (10)} & \textcolor{black}{D (16)} & \textcolor{black}{E (18)} & \textcolor{black}{F (20)} & x (11) & x (14) & x (15) \\
        \hline 
        Flight class & 3 & 3 & 3 & 3 & 3 &3 & 3 & 1 & 2 \\ 
        Fault mode & 1 & 1 & 1 & 2 & 2 & 2 & 2 & 2 & 2   \\
        Number of flights & 75 & 89 & 82 & 63 & 71 & 66 & 59 & 76 & 67  \\ 
          Mean length of flight (sec.) & 11,375 & 11,611 & 11,618 & 12,148 & 12.545 & 11,639 & 11,246 & 2,063 & 6,470 \\
          \hline 
     \end{tabular}
    \label{tab:info_engines}
\end{table*}

{\color{black}The N-CMAPSS dataset contains the measurements of 28 different sensors. However, many of the measurements of these sensors are highly correlated. Following \cite{de2023developing}, a total of 13 sensors (see Table \ref{tab:selected}) are selected  as input for the RUL prognostic model based on the correlation.}
{\color{black}
After this sensor selection, 89 million sensor measurements are still available. 
Previous studies using N-CMAPSS \cite{de2023developing, chao2022fusing} aggregate the measurements to reduce the computational time to train a neural network. 
Similarly, as measurement aggregation, the mean measurement per sensor and per operating condition per 20 seconds is considered in this study. This reduces the dataset to over 4 million measurements.} 

\textcolor{black}{Lastly, the measurements are normalized using min-max normalization. The measurements of a training engine (i.\,e., 2, 5, 10, 16, 18, 20) are normalized  using the minimum and maximum measurement belonging to only this engine. In this way, airlines do not share their minimum and maximum measurement with each other. The measurements of the test engines (i.\,e., 11, 14, 15), however, are normalized with the minimum and maximum measurement from all training engines (i.\,e., 2, 5, 10, 16, 18, 20), since these engines are not assigned to a specific airline.}

\begin{figure}[!ht]
    \centering
    \begin{subfigure}{0.37\textwidth}       
        \includegraphics[width = \textwidth]{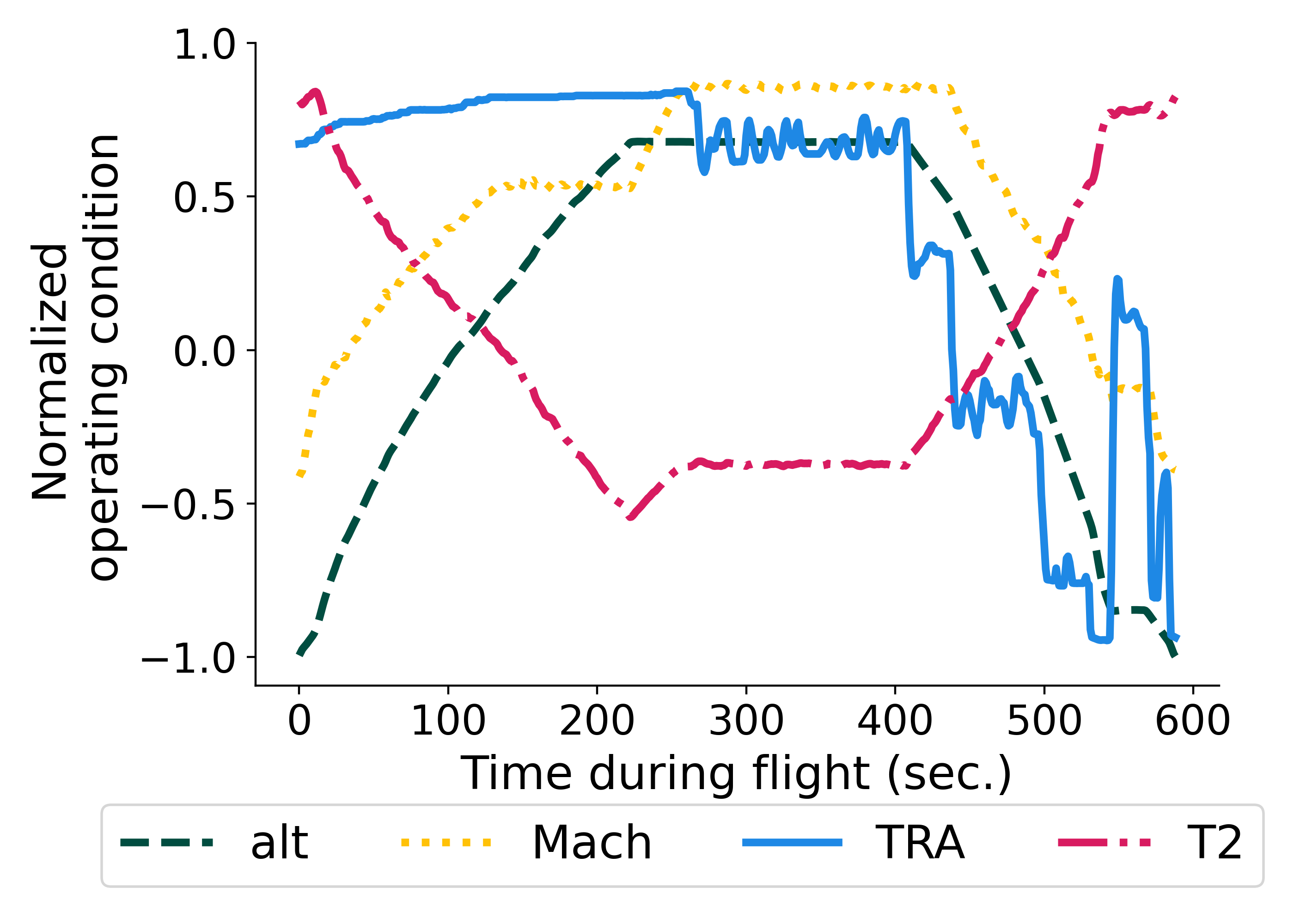}
        \caption{Normalized operating conditions.}
        \label{fig:operating}
    \end{subfigure}
    \begin{subfigure}{0.37\textwidth}       
        \includegraphics[width = \textwidth]{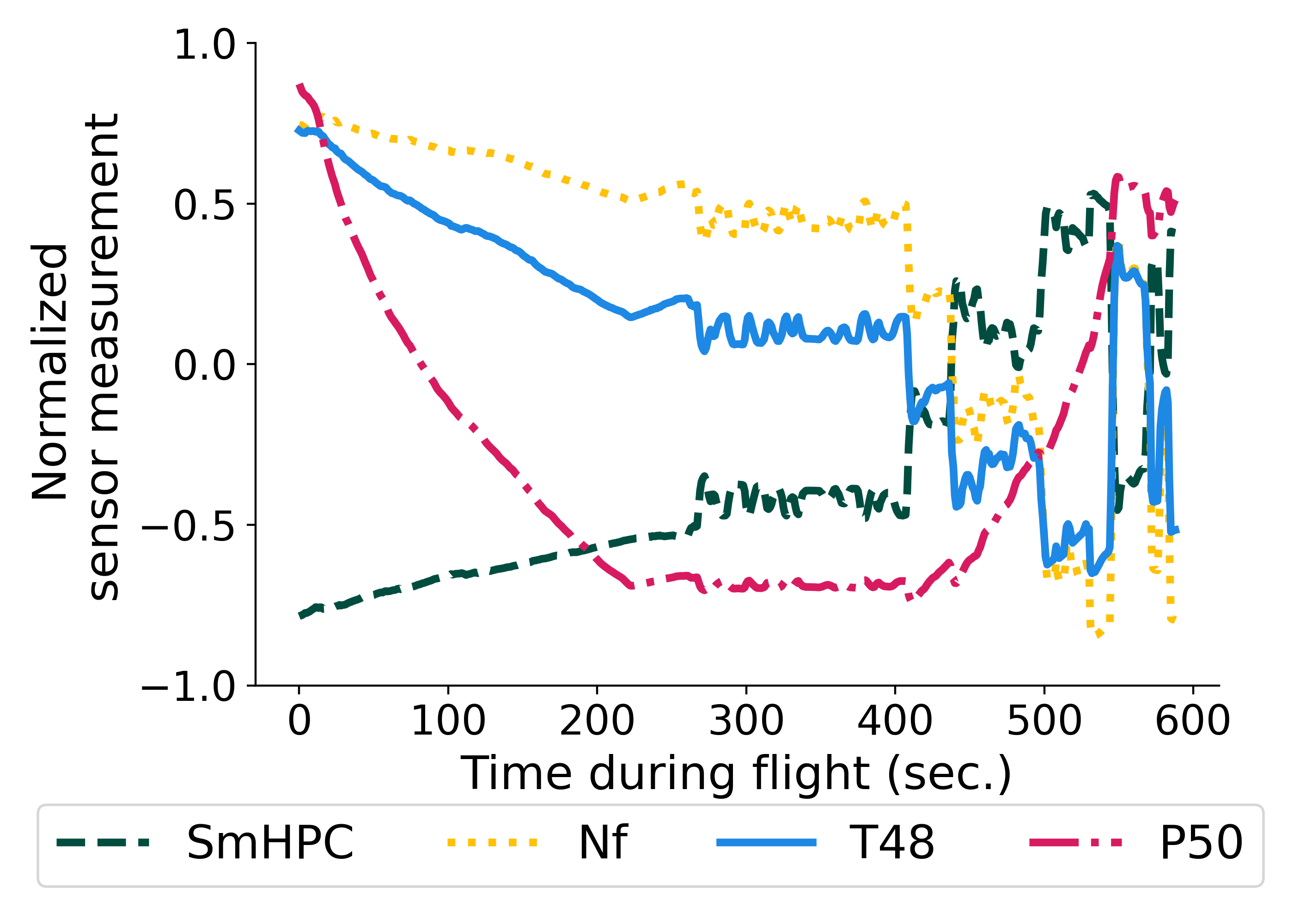}
        \caption{Normalized sensor measurements.}
        \label{fig:measurements}
    \end{subfigure}
    \caption{Normalized operating conditions and normalized sensor measurements of flight 1, training engine 2.}
    \label{fig:normalized_one_flight}
\end{figure}




{\color{black}
After normalization, a validation/training data split is performed for each training engine considered in the dataset: 20\% of the flights of each training engine are randomly selected as validation data, and the remaining 80\% of the flights of this engine become training data.
}

\textcolor{black}{
Figure \ref{fig:normalized_one_flight} shows an example of the considered  measurements. Figure \ref{fig:operating} shows the aggregated, normalized measurements of the four operating conditions during the first flight of training engine 2. Figure \ref{fig:measurements} shows the aggregated normalized measurements of four sensors during the same flight. Due to the min-max normalization, all measurements are between -1 and 1. The operating conditions and sensor measurements clearly show the flight profile. For instance, the altitude of the aircraft (``alt'') first increases during the take-off, is then stable during  cruise, and decreases when the aircraft is landing. Similarly, the pressure at the Low Pressure Turbine outlet (``P50'') is high during the take-off and landing, and low during cruise.}

\textcolor{black}{ Let $\mathbf{X}^{e,f} = \left[\mathbf{X}^{e,f}_1, \mathbf{X}^{e,f}_2, \ldots, \mathbf{X}^{e,f}_{M^{e,f}}\right]$ denote the normalized multi-sensor measurements during flight $f$ for an engine $e$. 
Here, $M^{e,f}$ is the number of measurements from this flight.  Each normalized multi-sensor measurement $\mathbf{X}^{e,f}_g$ (with $g \in \{1,2,\ldots,M^{e,f} \}$) contains  $H$ measurements, i.e., $\mathbf{X}^{e,f}_g = \left[x^{e,f,1}_g, x^{e,f,2}_g, \ldots, x^{e,f,H}_g\right]$. Also, $x^{e,f,h}_g$ (with $h \in \{1,2,\ldots,H\}$) is the $g^{\text{th}}$ normalized measurement of sensor $h$ during flight $f$ of engine $e$. Since 13 sensors and 4 operating conditions are considered in this case study, there are $H=17$ different measurements in total.} 

\textcolor{black}{
The number of measurements $M^{e,f}$ varies per engine $e$ and per flight $f$. The RUL prognostic model, however, can only handle input samples of a fixed dimension. To address this, for each engine $e$ and each flight $f$, several samples of length $50$ are extracted from $\mathbf{X}^{e,f}$. The first sample $\mathbf{\tilde{X}}^{e,f}_1$ of flight $f$ of engine $e$ consists of the first $50$ multi-sensor measurements of $\mathbf{X}^{e,f}$, i.e., $\mathbf{\tilde{X}}^{e,f}_1 = \left[\mathbf{X}^{e,f}_1, \mathbf{X}^{e,f}_{2}, \ldots, \mathbf{X}^{e,f}_{50}\right]$. Moving forward with a step size of $10$ time-steps, the second sample $\mathbf{\tilde{X}}^{e,f}_{11}$ of length $50$ from this flight is created as $\mathbf{\tilde{X}}^{e,f}_{11} = \Big[\mathbf{X}^{e,f}_{11}, \mathbf{X}^{e,f}_{12},$ $ \ldots, \mathbf{X}^{e,f}_{60}\Big]$. Again moving  $10$ time-steps, the third sample  $\mathbf{\tilde{X}}^{e,f}_{21}$ of this flight is created. This is repeated until the end of the flight. The data sample of engine $e$, flight $f$ and time-step $s$ is denoted by $\mathbf{\tilde{X}}^{e,f}_s = [\mathbf{X}^{e,f}_s, \mathbf{X}^{e,f}_{s+1}, \ldots, \mathbf{X}^{e,f}_{s + 50 - 1}]$. Let $S^{e,f}$ be the set with all the time-steps $s$ belonging to flight $f$ of engine $e$.  }

\subsection{Convolutional Neural Network for RUL prognostics} \label{sec:cnn}

\begin{figure*}[!ht]
    \centering
    \includegraphics[width = 0.75\textwidth]{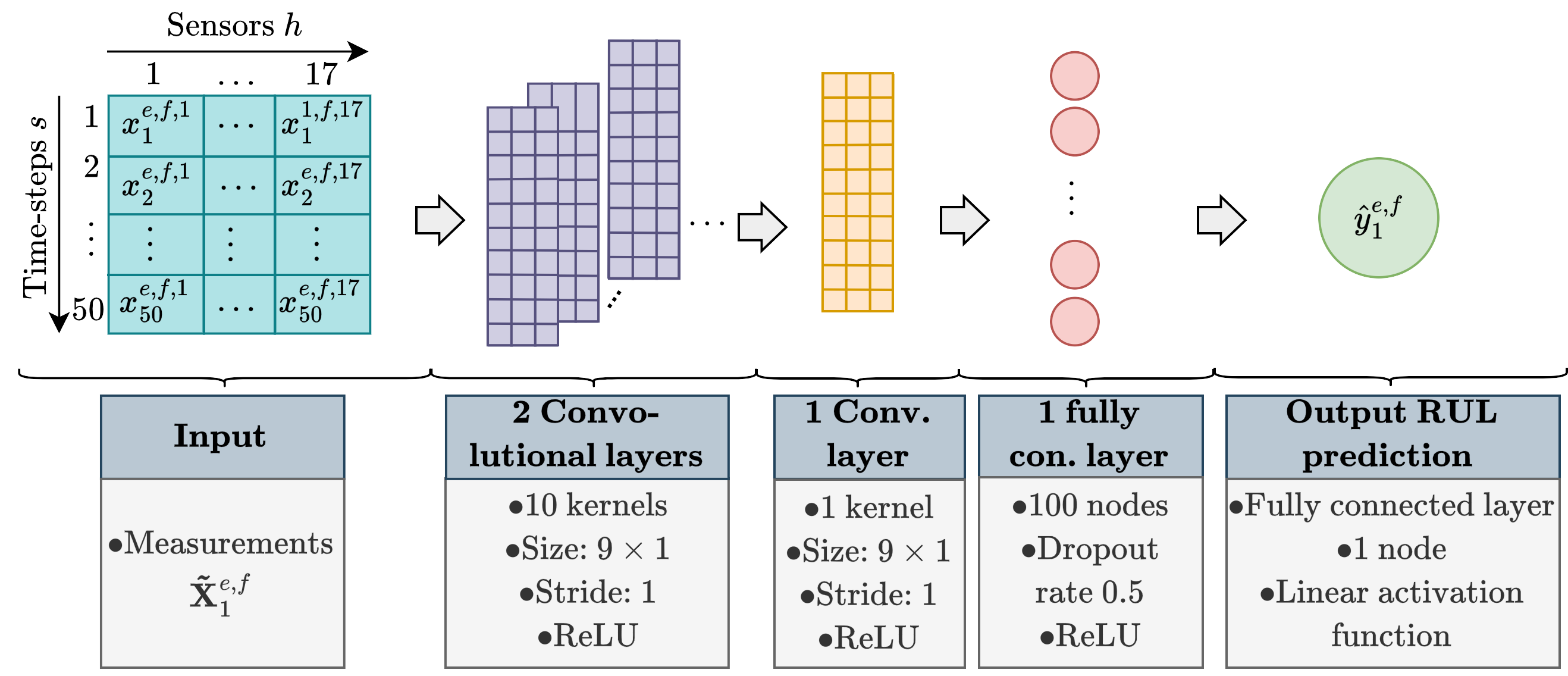}
    \caption{Schematic overview of the CNN with a data sample $\mathbf{\tilde{X}}^{e,f}_1$.}
    \label{fig:cnn}
\end{figure*}

\textcolor{black}{The RUL of the aircraft engines is predicted using an one-dimensional Convolutional Neural Network (1D-CNN). CNNs are originally developed for image data. However, the ability of CNNs to extract features from the data makes them also very suitable for RUL prediction from time-series of sensor measurements \cite{li2018remaining}. Especially 1D-CNNs have been frequently used for RUL prediction using multi-variate time series, where their ability to extract features yielded state-of-the-art results  \cite{li2018remaining, wang2019deep, shang2024novel}. }

Figure \ref{fig:cnn} shows a schematic overview of the considered CNN. The CNN consists of $3$ convolutional layers. The first $2$ convolutional layers consist of $10$ kernels, each of size $9 \times 1$ (i.e., one-dimensional kernels), that move with a stride of 1. The last convolutional layer has  one kernel, and compresses all $10$ feature maps into one single feature map. The convolutional operation in the $c^{th}$ convolutional layer for the $p^{th}$ kernel  $k^c_p$ is \cite{wang2019deep}:
\begin{equation}
    z^c_p = \text{ReLU}(k^c_p * z^{c-1} + b^c_p)
\end{equation}
where $z^c_p$ is the $p^{th}$ feature map of layer $c$, with $b^c_p$ the corresponding bias of this feature map, $z^{c-1}$ represents all feature maps in layer $c-1$, * is the convolutional operator and $\text{ReLU}(\cdot)$ is the considered activation function, namely the Rectified Linear Unit.

After the convolutional layers, the CNN consists of two fully connected layers. The first fully connected layer consists of $100$ nodes. Let $z^C$ be the feature map from the last convolutional layer $C$. The output $z^l$ of the first fully connected layer is then \cite{wang2019deep}:
\begin{equation}
    z^l = \text{ReLU}(w^l z^{C} + b^l),
\end{equation}
with $b^l$ the bias and $w^l$ the weights of the fully connected layer. A dropout rate of 0.5 is applied on the output of this layer to prevent overfitting.  Lastly, the final layer contains  one node, and outputs the RUL prognostic using a linear activation function. \textcolor{black}{The  CNN is trained with the Adam optimizer \cite{kingma2014adam} for 100 epochs with a batch size of 128 and a learning rate of 0.001. }

The Root Mean Squared Error (RMSE) between the true RUL and the estimated RUL is used as the loss function to train this CNN. With the new aggregation methods, the RMSE is also used as the metric loss $\mathcal{L}^{\text{metric}}(V_i, W_j)$ to compute the evaluation score $E_j$ of a local model of an airline $j \in C$ in Section \ref{sec:allocation}. To validate the global model, with the parameters $W_G$ (Step 5 in the FL framework, Section \ref{sec:fed_valid}), the sum of the loss is computed for each airline $i \in C$ as the sum of the squared errors: 
\begin{equation}
    \mathcal{L}^{\text{sum}}(V_i, W_G) = \sum_{j = 1}^{|V_i|} \left(\hat{y}_j - y_j\right)^2,
\end{equation}
 where $y_j$ is the true RUL of the $j^{\text{th}}$ sample in the validation set $V_i$, and $\hat{y}_j$ is the estimated RUL of this sample. 

\textcolor{black}{As described in Section \ref{sec:data}, the data of each flight $f$ of engine $e$ is divided into multiple data samples, where each data sample $\mathbf{\tilde{X}}^{e,f}_s$ belongs to a time-step $s$ of flight $f$ of engine $e$. The set $S^{e,f}$ contains all time-steps $s$ belonging to flight $f$ of engine $e$.  Each data sample $\mathbf{\tilde{X}}^{e,f}_s, s \in S^{e,f}$ is used as input to the CNN, to obtain as output the predicted RUL $\hat{y}^{e,f}_s$. The final estimated RUL $\hat{y}^{e,f}$ of engine $e$ after flight $f$ is the median of the RUL prognostic of all samples belonging to that flight:
\begin{equation}
    \hat{y}^{e,f} = \text{Median}_{s \in S^{e,f}}\left(\hat{y}^{e,f}_s\right).
\end{equation}
}

\subsection{Benchmark models} 
\label{sec:benchmark}

The FL framework is compared with two benchmark models: the Unrestricted access Centralized (UC) learning model and the Non-collaborative Isolated (NI) learning model. 

\subsubsection*{Unrestricted access Centralized (UC) learning model } \label{sec:uc_model_case}

For this model, it is envisioned that all airlines fully cooperate by sharing all their data, without any privacy barriers. \textcolor{black}{A central server trains a RUL prognostic model on engines 2, 5, 10, 16, 18 and 20 and tests this model on engines 11, 14 and 15.
}This is an idealistic case when all data is shared. The outcome of this benchmark is a lower bound on the performance of the FL framework.

For the UC learning model, the CNN is trained with all the available  data of all six training engines. \textcolor{black}{The sensor measurements and operating conditions are normalized using min-max normalization. Since all data is available at a central server, the minima and maxima across all data from all training engines is used for normalization. The training/validation split after normalization remains the same as in the FL framework. 
} 

\subsubsection*{Non-collaborative Isolated (NI) learning model} \label{sec:ni_learning_case}



\begin{table*}[!ht]
    \centering
    \caption{Comparison of the RMSE and MAE (both in flights) of the FL model and the Unrestricted Access centralized (UC) learning model. The lowest obtained RMSE for the validation set, and the number of epochs after which this lowest RMSE is obtained, are also given. The best results are denoted in bold.  }
    \begin{tabular}{c|cc|cccccc|cc}
    & \multicolumn{2}{c|}{Validation results} & \multicolumn{8}{c}{Test results}\\
    \cline{2-11} 
    &  Min. valid. & Obt. after & \multicolumn{2}{c|}{Engine 11} & \multicolumn{2}{c|}{Engine 14} & \multicolumn{2}{c|}{Engine 15} & \multicolumn{2}{c}{All engines}  \\ 
   Model & RMSE & ... epochs &  RMSE & \multicolumn{1}{c|}{MAE} & RMSE & \multicolumn{1}{c|}{MAE}  & RMSE & \multicolumn{1}{c|}{MAE}  & RMSE & MAE     \\ 
     \hline
     FL model & 12.7  & 61 & \textbf{6.2} & \textbf{5.0} & 13.5 & 9.2 & 7.4 & 5.3 & 9.9 & 6.7  \\   
    UC model& \textbf{9.8} & 85 &  6.3 & \textbf{5.0} & \textbf{8.0} & \textbf{5.6} & \textbf{3.7} & \textbf{2.4} & \textbf{6.3} & \textbf{4.4}  
    \end{tabular}
    \label{tab:predictions}
\end{table*}

{\color{black}
For the NI model, it is envisioned that there is no collaboration between airlines. Each airline trains its own RUL prognostic model with the data of the one  engine it has available. The considered cases are as follows: Airline A (engine 2) trains a CNN with the data of engine 2, Airline B (engine 5) trains a CNN with the data of engine 5,  Airline C (engine 10) trains a CNN with the data of engine 10,  Airline D (engine 16) trains a CNN with the data of engine 16,  Airline E (engine 18) trains a CNN with the data of engine 18 ,  and Airline F (engine 20) trains a CNN with the data of engine 20 (see Table \ref{tab:info_engines}). In total, six different CNNs are trained, one CNN per airline. As in the FL framework,  the measurements of a training engine are normalized  using the minimum and maximum measurement belonging to only this engine. After normalization, the training/validation split  remains the same as in the FL framework.}

\textcolor{black}{
The performance of the proposed FL framework vs. the NI learning model is analyzed by testing these six CNNs on the three test engines 11, 14 and 15. 
In contrast to the FL framework, when testing the CNN of airline $i, i 
\in \{A, B, C, D, E, F \}$, the measurements of the test engines are normalized with the minimum and maximum measurement from the one training engine belonging to airline $i,  i
\in \{A, B, C, D, E, F \}$. 
}

\subsection{Simulation of 
noisy condition-monitoring data} \label{sec:noise_case}

In Section \ref{sec:method}, four novel aggregation methods are proposed to develop a global model that is resilient to clients with noisy data. 
To evaluate these aggregation methods in the presence of noisy data,  a scenario is simulated where some airlines have noisy data. Here,  a varying amount of noise is added to both the sensor measurements and the operating conditions of the engine assigned to these airlines. 

Assume that  engine $e_i$ is available at airline $i \in C$, and that this airline $i$ is selected such that noise is added to the measurements.  
  Noise is added \textit{before} normalizing the data. For each non-normalized measurement $\check{x}^{e_i,f,h}_{g}$ (i.e., the $g^{\text{th}}$  non-normalized measurement of sensor $h$ during flight $f$ of engine $e_i$), noise is sampled from a normal distribution with a mean of $0$ and a standard deviation of $\alpha \cdot \sigma_{i,h}$: 
\begin{align}
    & \check{x}^{e_i,f,h}_{g} \leftarrow \check{x}^{e_i,f,h}_{g} + \mathcal{N}(0,\, \alpha \cdot \sigma_{i,h}),  \\ 
   & \forall f \in \{1,\ldots,F_{e_i}\},  \forall g \in \{1,\ldots,M^{e_i,f}\}, \forall h \in \{1,\ldots,H\}. \nonumber 
    \label{eq:sensor_noise}
\end{align}
Here, $\alpha$ is an user-chosen parameter, $F_{e_i}$ is the number of flights of engine $e_i$ and $\sigma_{i,h}$ is the standard deviation of all non-normalized measurements of a sensor $h$ for all flights of the considered engine $e_i$.

\section{Results} 
\label{sec:results}


\subsection{Results for the FL (Federated Learning) model vs the UC learning (centralized) model}  \label{sec:results_pd_uc}

 \begin{figure}[!ht]
    \centering
    \includegraphics[width = 0.43\textwidth]{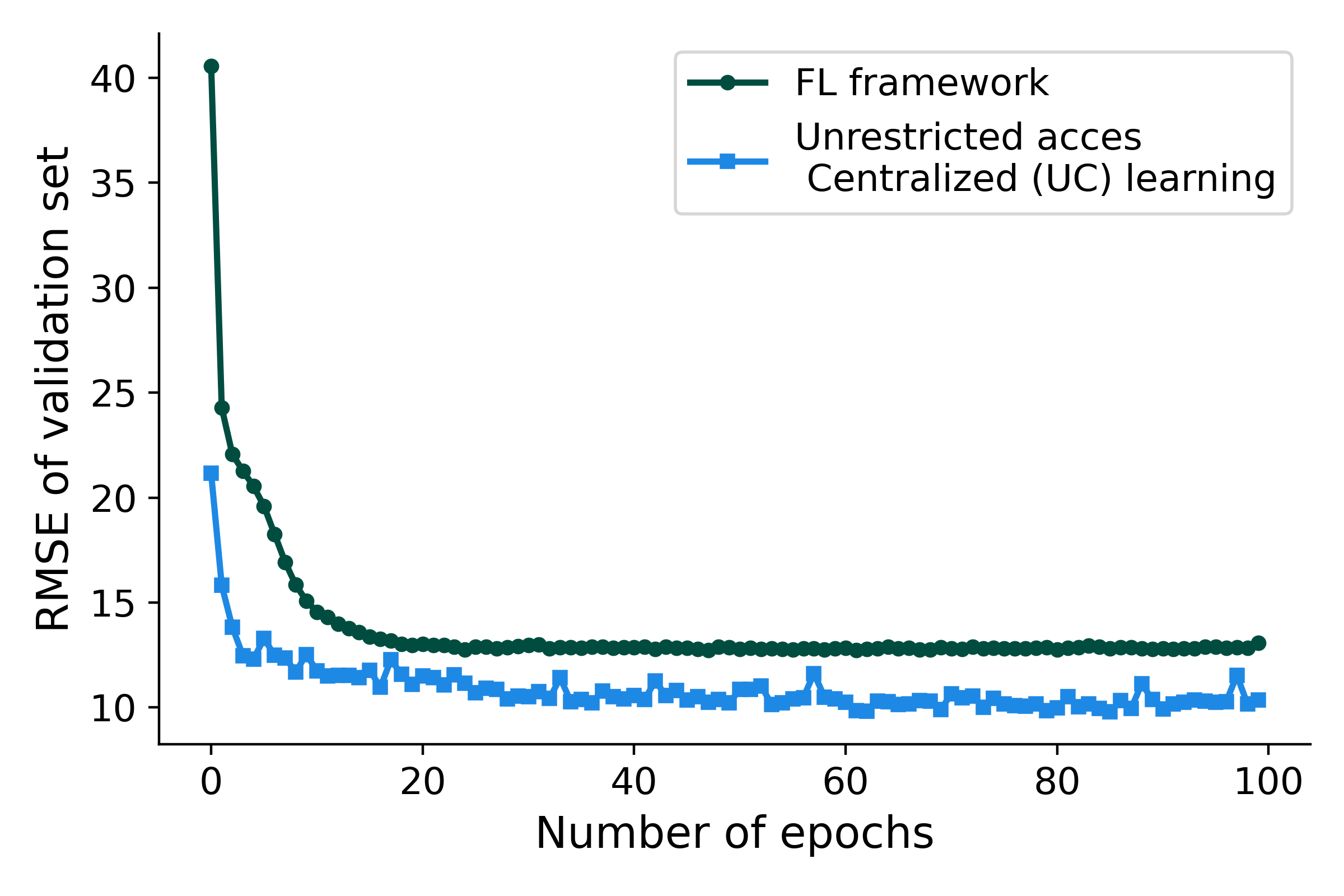}
    \caption{RMSE of the validation set for the FL model and the unrestricted access centralized (UC) learning model.}
    \label{fig:valid_dis_cen}
\end{figure}

\begin{figure}[!ht]
    \centering
    \begin{subfigure}{0.28\textwidth}       
        \includegraphics[width = \textwidth]{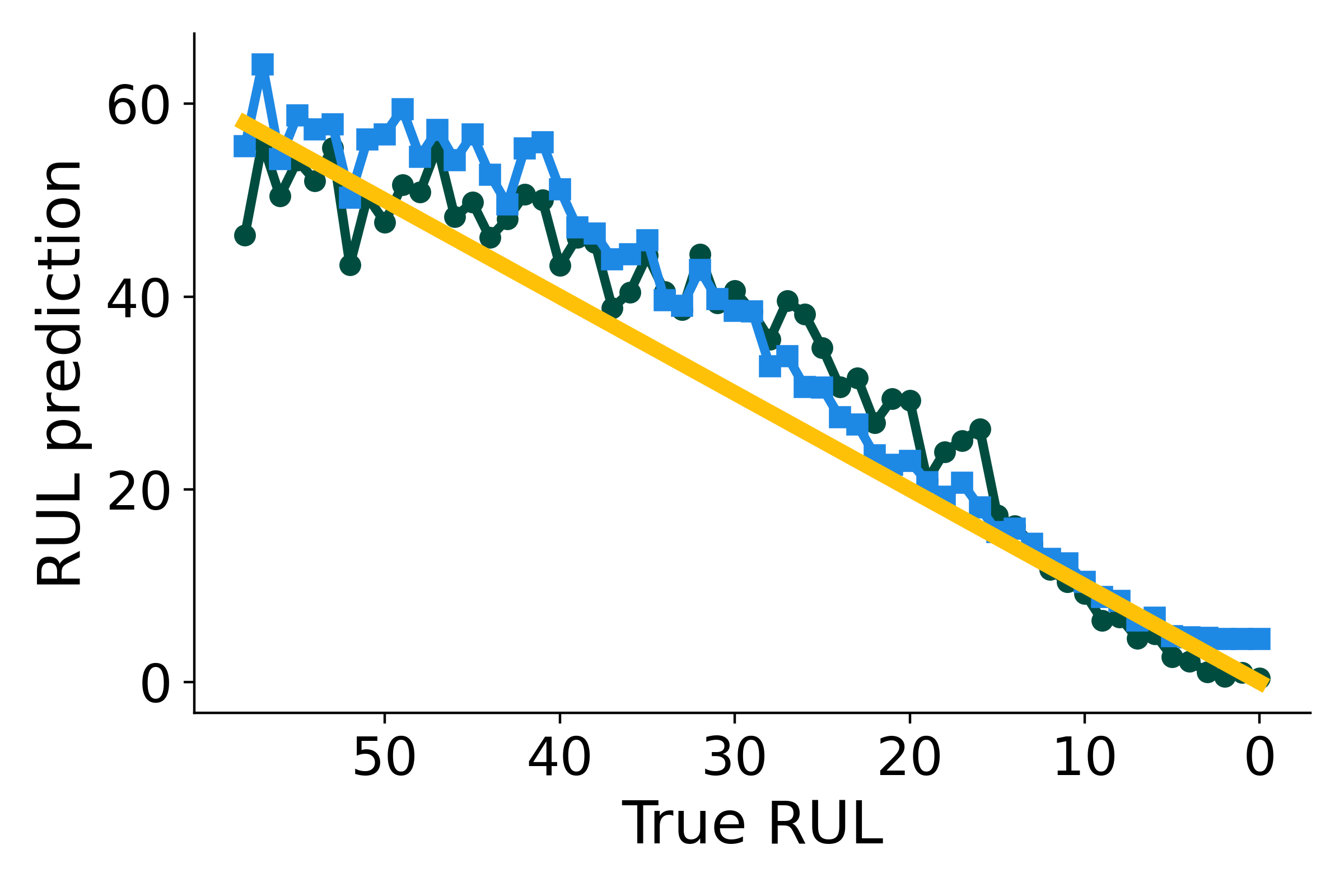}
         \caption{Engine 11}
        \label{fig:pred_11}
    \end{subfigure}
    \begin{subfigure}{0.28\textwidth}       
        \includegraphics[width = \textwidth]{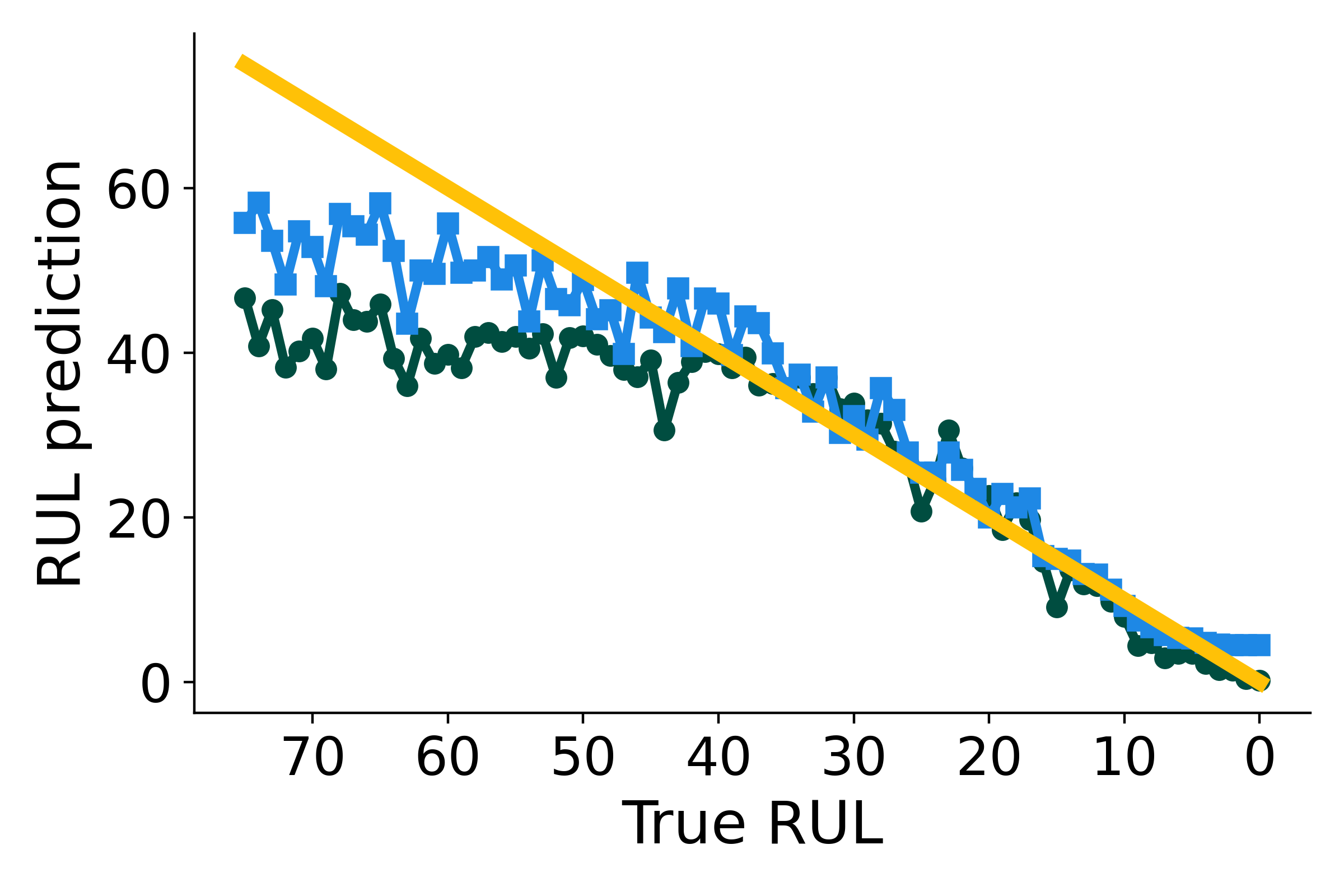}
         \caption{Engine 14}
        \label{fig:pred_14}
    \end{subfigure}
    \begin{subfigure}{0.28\textwidth}       
        \includegraphics[width = \textwidth]{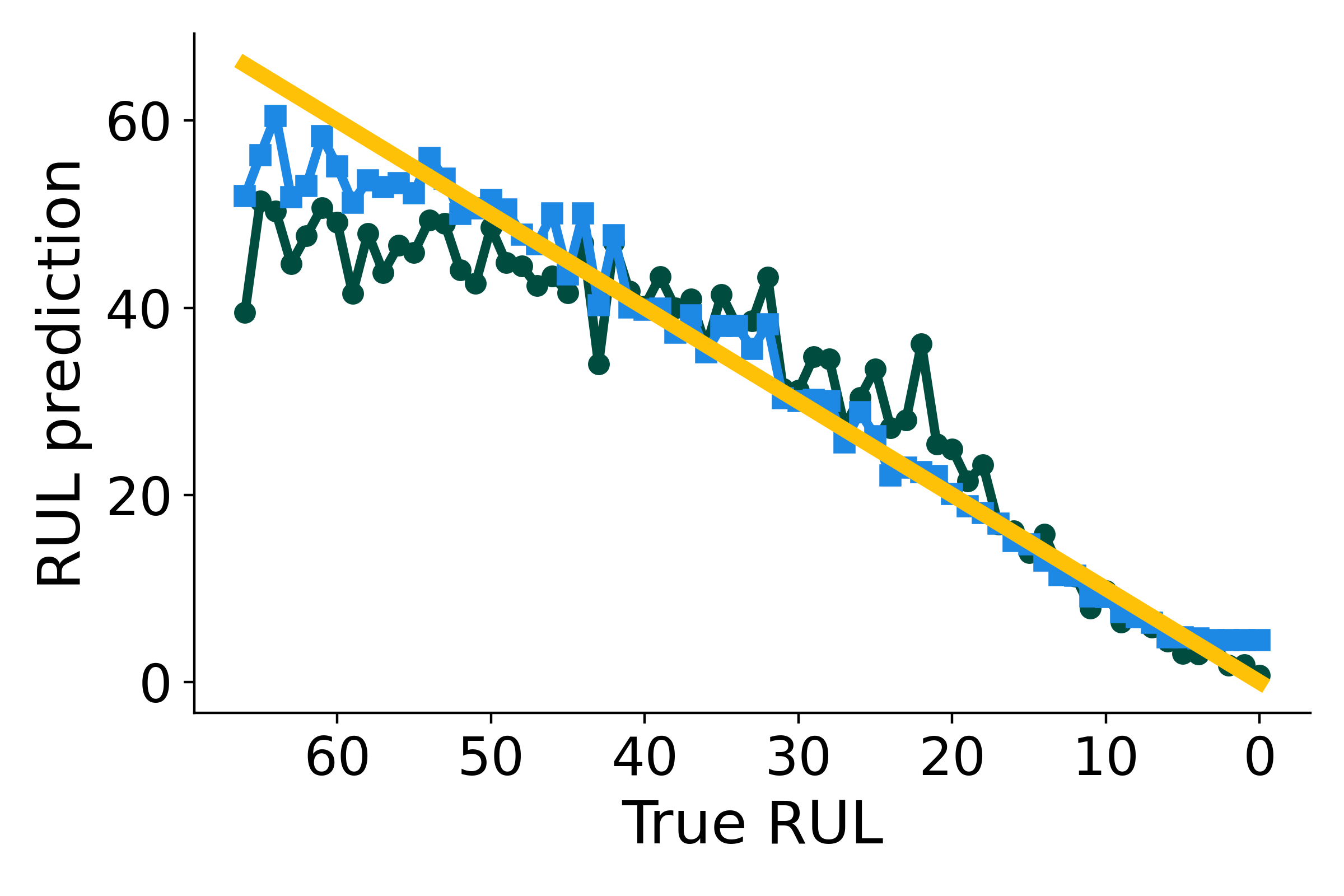}
           \caption{Engine 15}
        \label{fig:pred_15}
    \end{subfigure}
     \includegraphics[width = 0.33\textwidth]{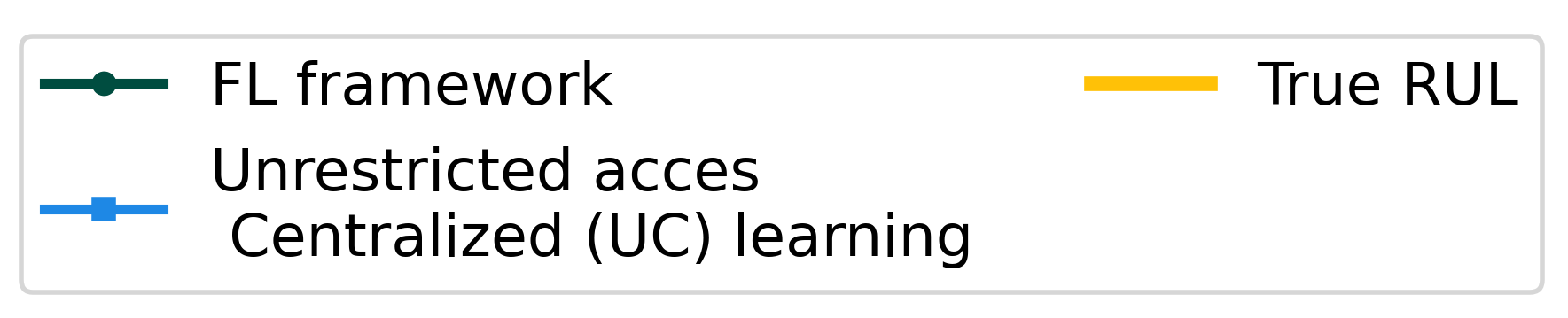} 
    \caption{RUL prognostics per engine for the FL model and the Unrestricted Access centralized (UC) learning model.}
    \label{fig:predictions}
\end{figure}

\begin{table*}[!ht]
    \centering
        \caption{  Comparison of the RMSE and MAE (both in flights) of the FL model and the six non-collaborative isolated (NI) learning models (each trained with only the training data of one engine). The best results are denoted in bold.  }
    \begin{tabular}{c|cccccc|cc} 
     & \multicolumn{8}{c}{Test results} \\
     \cline{1-9} 
        &  \multicolumn{2}{c|}{Engine 11} & \multicolumn{2}{c|}{Engine 14} & \multicolumn{2}{c|}{Engine 15} & \multicolumn{2}{c}{All engines}  \\ 
       & RMSE & \multicolumn{1}{c|}{MAE} & RMSE & \multicolumn{1}{c|}{MAE}   & RMSE & \multicolumn{1}{c|}{MAE}  & RMSE & MAE     \\ 
     \hline 
     \multicolumn{9}{c}{FL model} \\
     \hline 
       & \textbf{6.2} & \textbf{5.0} & 13.5 & 9.2  &  7.4 & \textbf{5.3} &  9.9 & \textbf{6.7}   \\ 
      \hline 
      \multicolumn{9}{c}{Six NI learning models} \\ 
      \hline 
      \textcolor{black}{Airline A (Engine 2)} & 22.9 & 20.0 &  22.3 & 19.1  & 21.1 & 19.2  & 22.1 & 19.4 \\ 
      \textcolor{black}{Airline B (Engine 5)} & 8.4 & 6.5 &16.3 & 11.6 & 10.1 & 7.0 & 12.4 & 8.6   \\ 
      \textcolor{black}{Airline C (Engine 10)} & 9.6 & 7.4 & 25.7 & 21.1 & 17.2 & 13.7 & 19.3 & 14.6 \\ 
      \textcolor{black}{Airline D (Engine 16)} & 6.8 & 5.6 & \textbf{10.8} & \textbf{8.0} & \textbf{7.0} & 5.6 & \textbf{8.6} & 6.9 \\ 
      \textcolor{black}{Airline E (Engine 18)} & 18.8 & 15.9 & 17.2 & 14.3 & 15.9 & 13.7 & 17.3 & 14.6 \\ 
      \textcolor{black}{Airline F (Engine 20)} & 10.6 & 8.9 & 16.9 & 13.3 & 16.4 & 13.8 & 15.2 & 12.2  \\ 
      \hline
      Mean over all NI learning models & 12.9 & 10.7 & 18.2 & 14.7 & 14.6& 12.2 & 15.8 & 12.7 
    \end{tabular}
    \label{tab:predictions_cen_vs_iso}
\end{table*}

In this section, the results achieved with the proposed FL model are presented.  
These results are compared with the unrestricted access centralized (UC) learning model, where the ideal case in which all airlines share their data to train a common RUL prognostic model is considered (see Section \ref{sec:uc_model_case}).

 Table \ref{tab:predictions} shows the minimum RMSE of the validation set for both models. 
 Figure \ref{fig:valid_dis_cen} shows the RMSE of the validation set after each epoch. The RMSE of the validation set is consistently lower for the UC learning model, than for the FL model. This is  expected, since the centralized model is directly trained with all the data from all the airlines.  For the FL model, the minimum validation loss is already obtained after 61 epochs, while  the model is trained for 100 epochs. At the end of the training procedure, the model might thus overfit. However, with the decentralized validation procedure, the weights of the model after 61 epochs are identified as the best weights to generate the test results with.


 Table \ref{tab:predictions} also shows the RMSE and Mean Absolute Error (MAE) of the RUL prognostics for the three test engines, while Figure \ref{fig:predictions} shows the corresponding RUL prognostic over time.  Also here, the UC learning model outperforms the FL model, with an equal or lower RMSE and MAE of the RUL prognostics for each test engine, except the RMSE of test engine 11. However, the overall RMSE of the FL model (9.9 flights) is close to the overall RMSE of the UC learning model (6.3 flights). 

\subsection{Results for the FL model vs the NI learning (isolated) models} \label{sec:results_pd_ni}

In this section, the results achieved with the proposed FL model are compared with the results achieved with the non-collaborative (NI) learning models. \textcolor{black}{The aim of this comparison is to determine whether a better RUL prognostic model is obtained if airlines collaborate through the FL framework instead of each training their own prognostic model (NI learning, see Section \ref{sec:ni_learning_case}).}

Table \ref{tab:predictions_cen_vs_iso} shows the RMSE and MAE of the RUL prognostics for the test engines. Here, the FL model  has a lower overall RMSE (9.9 flights) and MAE (6.6 flights) than the mean RMSE (15.8 flights) and mean MAE (12.7 flights) over all six NI learning models. Moreover,  the FL model consistently outperforms five out of the six NI learning models, with a lower RMSE and MAE for all three test engines. 

However, the overall RMSE of the model of airline D, trained with engine 16, is 8.6 flights, which is lower than the overall RMSE of the FL model (9.9 flights). 
 \textcolor{black}{This illustrates the heterogeneity in the experimental setup, where Airline D possesses an engine that has similar failure conditions and fault modes as those observed in the test dataset (see Table \ref{tab:predictions_cen_vs_iso}), while Airlines A, B, C, E and F do not benefit from such a representative dataset.}
However, an airline does not know beforehand which types of failures will occur in the future. As the FL model provides the best results for five out of six airlines,  an airline is likely to obtain better results when participating in the FL model. 
In conclusion, it is therefore indeed beneficial for an airline to train a common RUL prognostic model through the proposed FL framework.

\subsection{Comparison of the novel aggregation methods for the global model parameters} \label{sec:results_agg}

\begin{figure}[!ht]
    \centering
    \begin{subfigure}{0.38\textwidth}
         \includegraphics[width = \textwidth]{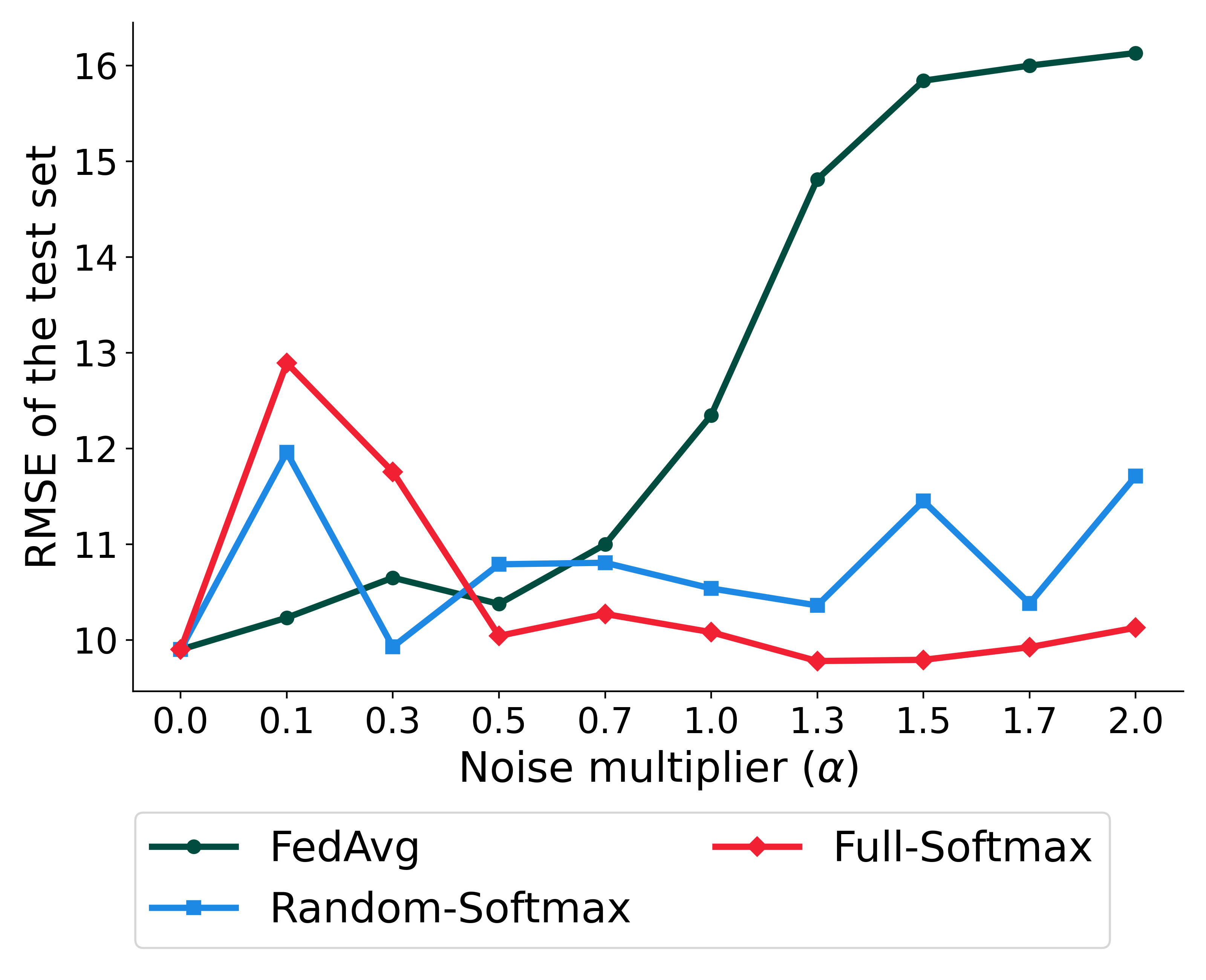}
        \caption{The RMSE of the test set with FedAvg, Random-Softmax and Full-Softmax. }
        \label{fig:rmse_noise_softmax} 
    \end{subfigure}
    \begin{subfigure}{0.38\textwidth}
         \includegraphics[width = \textwidth]{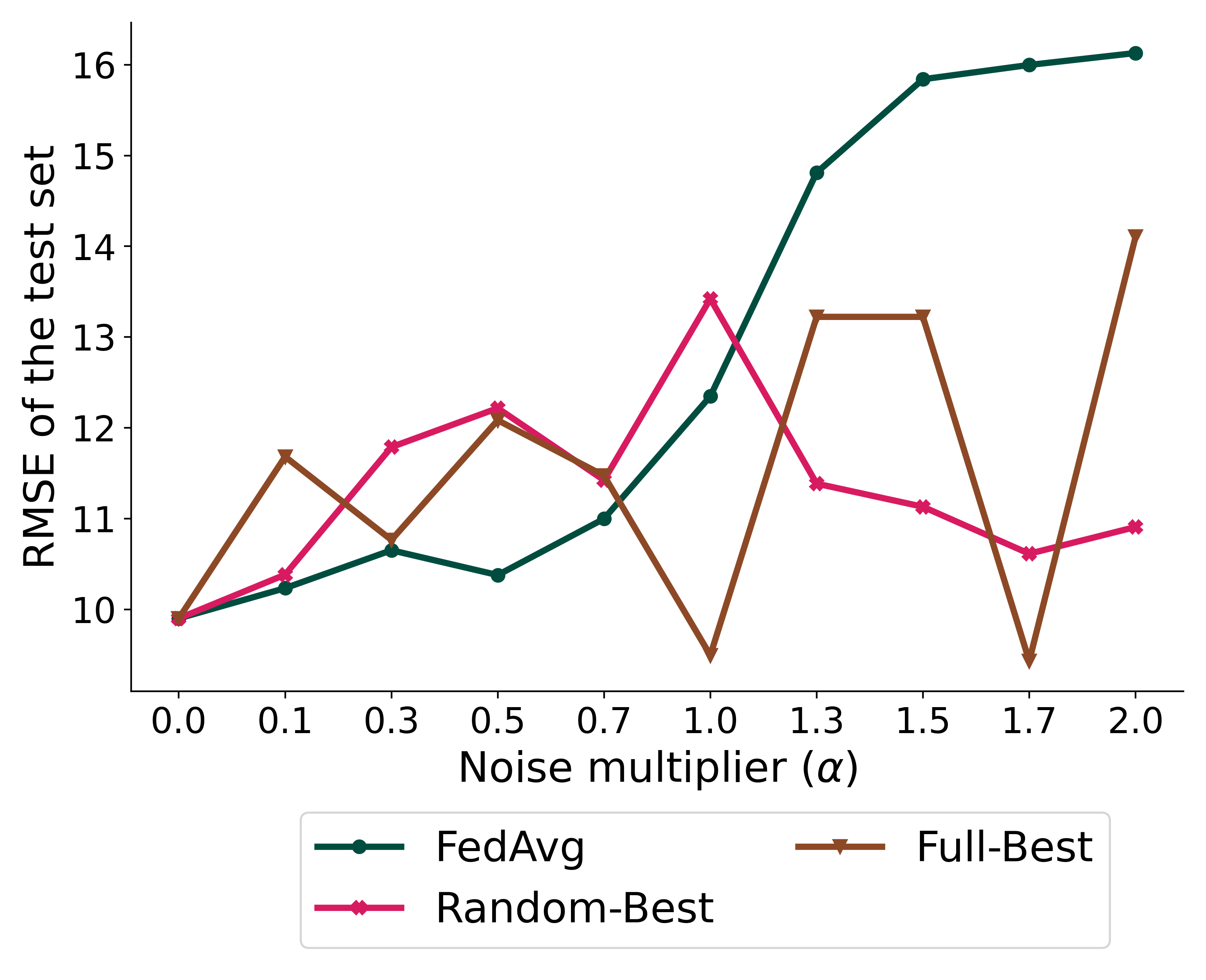}
        \caption{The RMSE of the test set with FedAvg, Random-Best and Full-Best. }
        \label{fig:rmse_noise_best} 
    \end{subfigure}
    \caption{The overall RMSE of the test set in flights for all considered aggregation methods, while considering different noise multipliers $\alpha$. Here, additional noise is added to the data of engine 5 and 18.  }
    \label{fig:rmse_noise_all}
\end{figure}

\begin{figure*}[!ht]
    \centering
    \begin{subfigure}{0.38\textwidth}       
        \includegraphics[width = \textwidth]{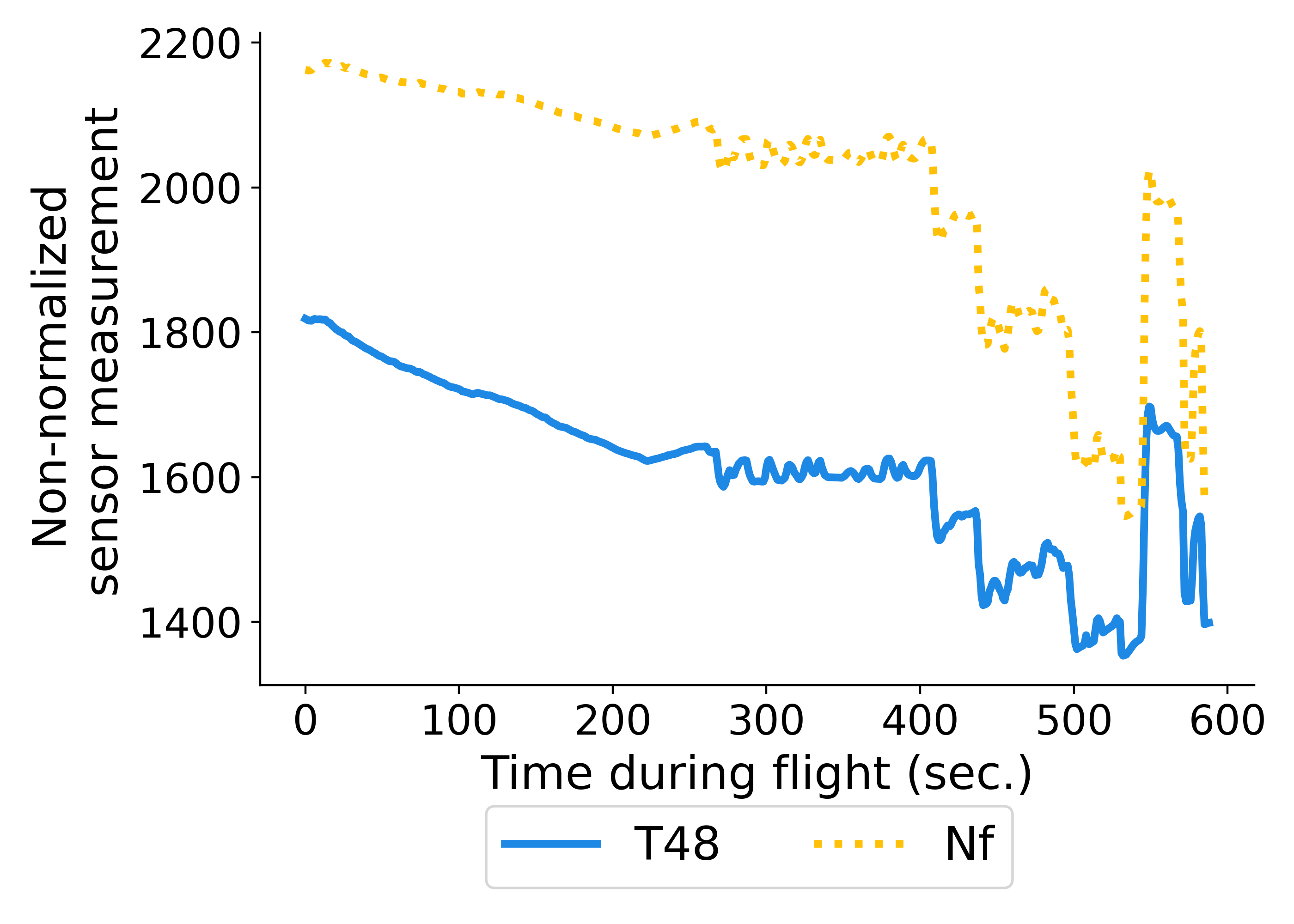}
        \caption{Non-normalized sensor measurements, $\alpha=0$.}
        \label{fig:measurements_alpha1}
    \end{subfigure}
        \begin{subfigure}{0.4\textwidth}       
        \includegraphics[width = \textwidth]{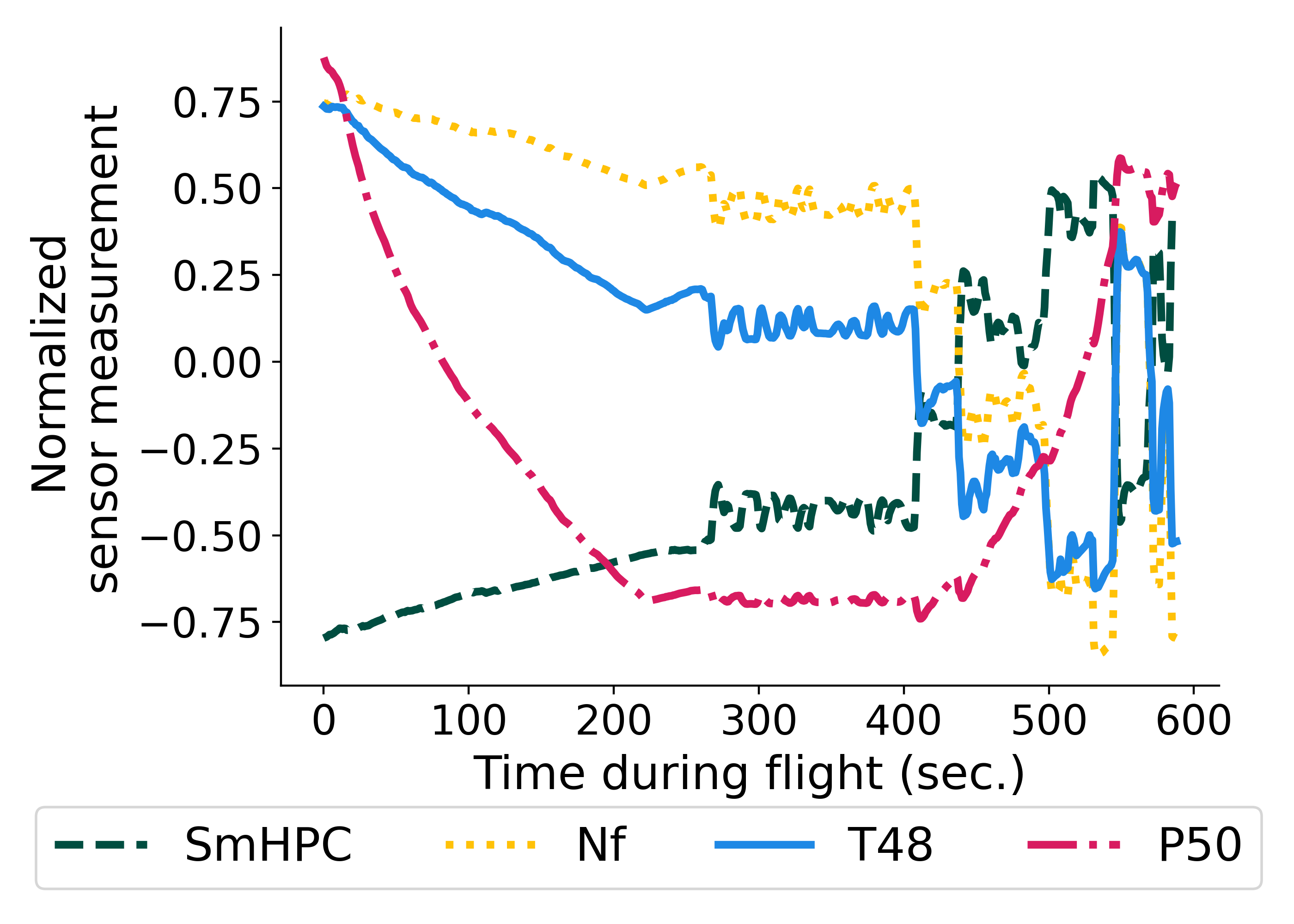}
        \caption{Normalized sensor measurements, $\alpha=0$.}
        \label{fig:measurements_alpha1}
    \end{subfigure}
        \begin{subfigure}{0.38\textwidth}       
        \includegraphics[width = \textwidth]{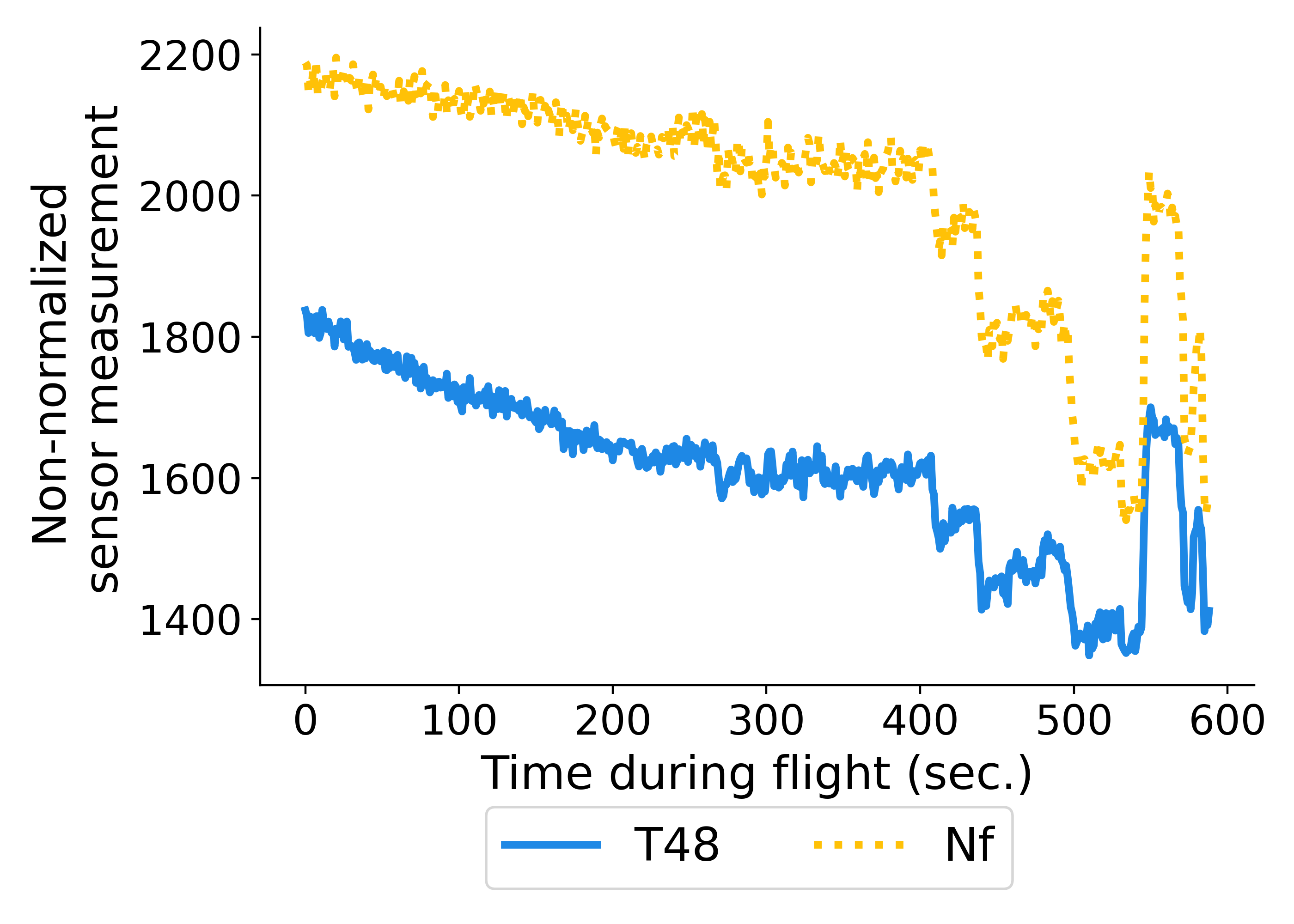}
        \caption{Non-normalized sensor measurements, $\alpha=0.1$.}
        \label{fig:measurements_alpha1}
    \end{subfigure}
    \begin{subfigure}{0.38\textwidth}       
        \includegraphics[width = \textwidth]{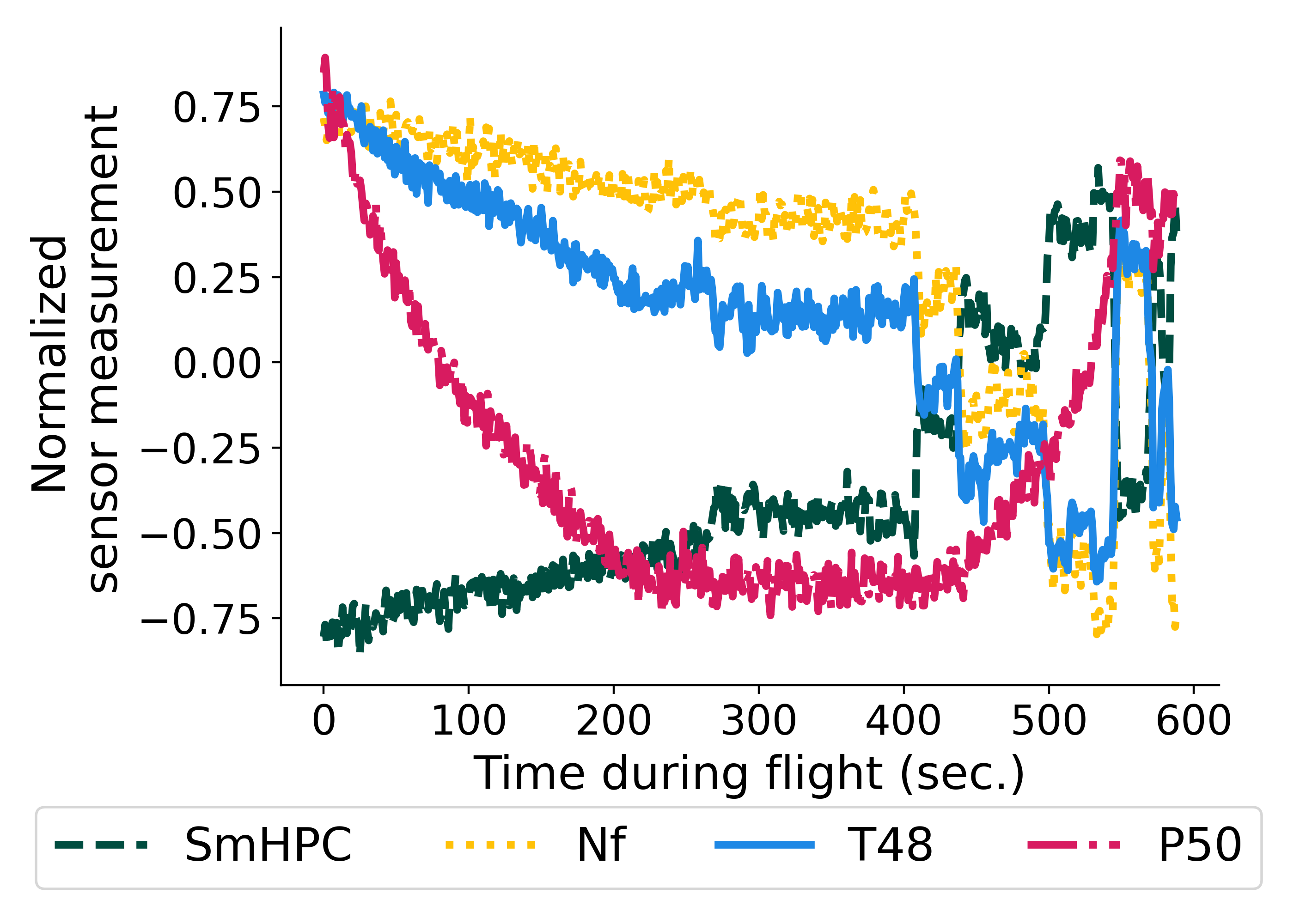}
        \caption{Normalized sensor measurements, $\alpha=0.1$.}
        \label{fig:measurements_alpha1}
    \end{subfigure}
    \begin{subfigure}{0.38\textwidth}       
        \includegraphics[width = \textwidth]{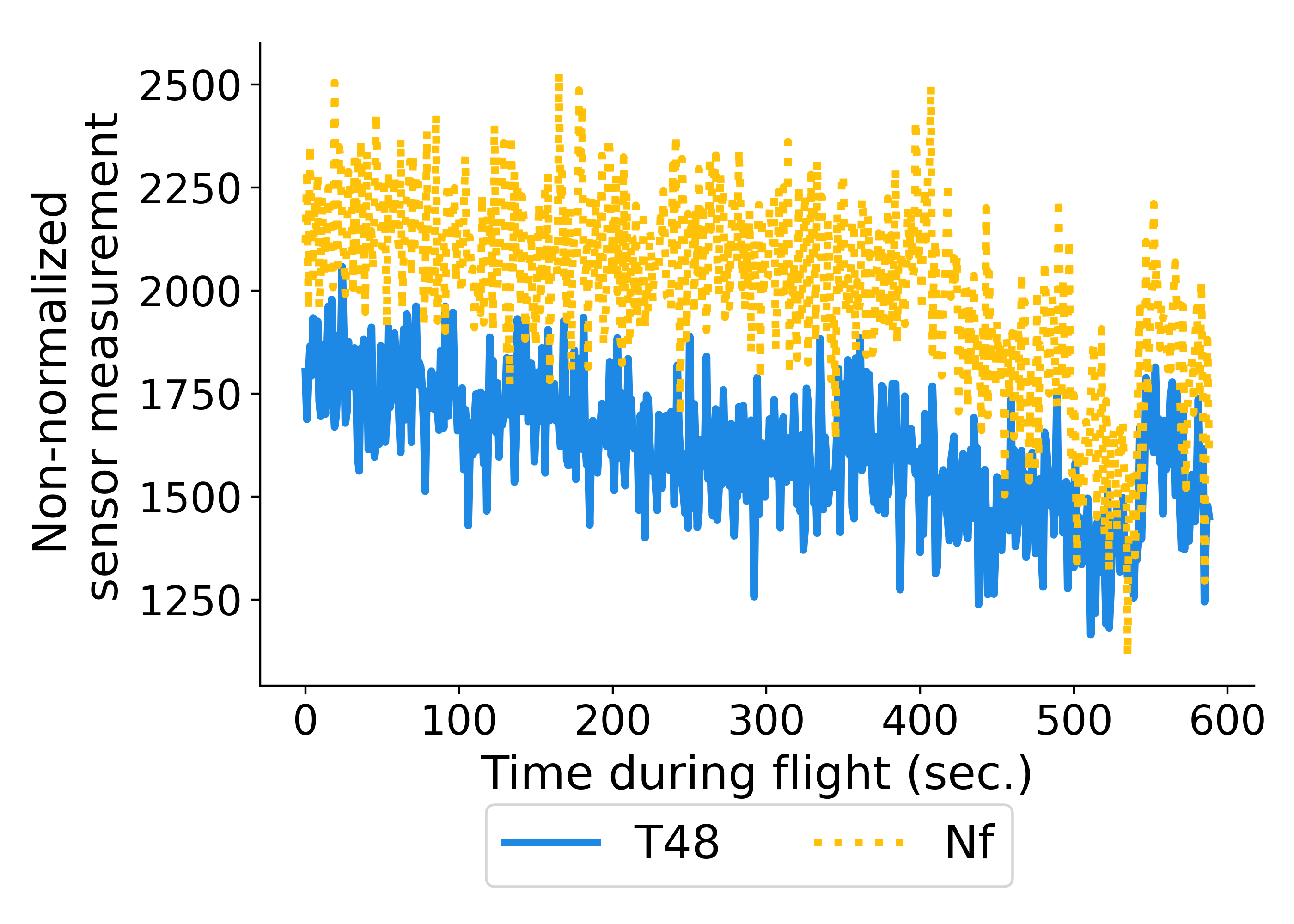}
        \caption{Non-normalized sensor measurements, $\alpha=1$.}
        \label{fig:measurements_alpha1}
    \end{subfigure}
    \begin{subfigure}{0.4\textwidth}       
        \includegraphics[width = \textwidth]{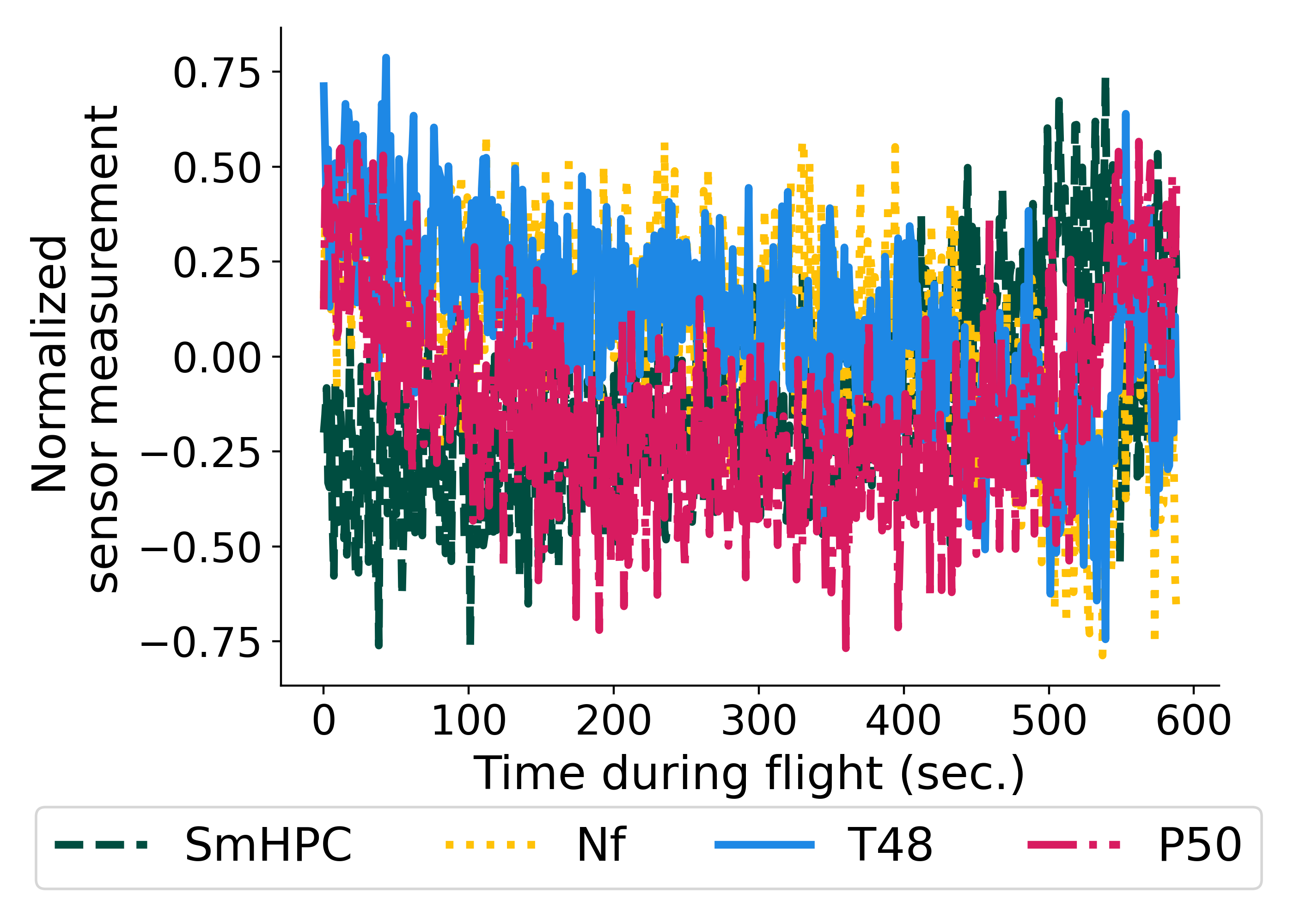}
        \caption{Normalized sensor measurements, $\alpha=1$.}
        \label{fig:measurements_alpha1}
    \end{subfigure}
    \caption{Non-normalized and normalized sensor measurements of flight 1, engine 2, for different values of the noise multiplier $\alpha$. For the non-normalized sensor measurements, only sensor T48 and Nf are plotted, since the scales of the other sensors are too different. }
    \label{fig:normalized_one_flight_alpha}
\end{figure*}

In this section, noise is added to the sensor measurements and operating conditions of training engine 5 and 18 (see Section \ref{sec:noise_case}).  Here, training engine 5 fails according to fault mode 1, while training engine 18 fails according to fault mode 2 (Section \ref{sec:case_study}). The size of the noise multiplier $\alpha$ is varied from 0 (no additional noise) to 2 (large additional noise). Next, the four proposed aggregation methods of Section \ref{sec:method} are compared with the standard FedAvg \cite{mcmahan2017communication} aggregation method (Section \ref{sec:fed_valid}).  

Figure \ref{fig:normalized_one_flight_alpha} shows a plot of both the normalized and the non-normalized measurements  of four selected sensors during flight 1, engine 2, with added noise with a multiplier of $\alpha = 0$, $\alpha=0.1$ and of $\alpha = 1$. 

Table \ref{tab:predictions_aggregation} shows the RMSE and MAE of the RUL prognostics of the three test engines, for all considered  aggregation methods, when there is no noise added to the data. The FedAvg method provides the best results, but the results with the softmax aggregation policy are only slightly worse. Full-Softmax and Random-Softmax have the same MAE for all test engines, but the RMSE of all test engines is considerable lower with Full-Softmax.  The best model aggregation policy (Random-Best and Full-Best) gives a relatively high RMSE and MAE.  

\begin{table*}[!ht]
    \centering
       \caption{Comparison of the RMSE and MAE (in flights) of the RUL prognostics for all three test engines in N-CMAPSS, for all considered aggregation methods. Here, no noise is added to the data (i.e., $\alpha = 0$).  The best results are denoted in bold.  }
    \begin{tabular}{c|cccccc|cc}
    Aggregation & \multicolumn{2}{c|}{Engine 11} & \multicolumn{2}{c|}{Engine 14} & \multicolumn{2}{c|}{Engine 15} & \multicolumn{2}{c}{All engines}  \\ 
   Method &   RMSE & \multicolumn{1}{c|}{MAE} & RMSE & \multicolumn{1}{c|}{MAE}  & RMSE & \multicolumn{1}{c|}{MAE}  & RMSE & MAE     \\ 
     \hline
     FedAvg  & \textbf{6.2} & \textbf{5.0} & 13.5 & \textbf{9.2} & \textbf{7.4} & \textbf{5.3} & \textbf{9.9} & \textbf{6.7}    \\ 
     \hline
     \multicolumn{9}{c}{Softmax aggregation policy} \\
     \hline
     Random-Softmax  & 8.6 & 7.0 & 15.1 & 10.2 & 7.7 & \textbf{5.6} & 11.2 & 7.7  \\
       Full-Softmax  & 8.3 & 6.9 & \textbf{12.6} & 9.3 & 8.1 & 6.6 & 10.1 & 7.7\\
      \hline
     \multicolumn{9}{c}{Best model aggregation policy} \\
     \hline
    Random-Best   & 11.2 & 9.4 & 13.6 & 12.0 & 11.3 & 9.9 & 12.2 & 10.6 \\  
     Full-Best & 8.1 & 6.5 & 15.2 & 10.6 & 9.2 & 6.8 & 11.6 & 8.1  
    \end{tabular} 
    \label{tab:predictions_aggregation}
\end{table*}

\begin{table*}[!ht]
    \centering
     \caption{Comparison of the RMSE and MAE (in flights) of the RUL prognostics for all three test engines in N-CMAPSS, for all considered aggregation methods. Here, additional noise with a standard deviation multiplier of $\alpha =0.1$ is added to the data of engine 5 and 18. The best results are denoted in bold. }
    \begin{tabular}{c|cccccc|cc}
   Aggregation & \multicolumn{2}{c|}{Engine 11} & \multicolumn{2}{c|}{Engine 14} & \multicolumn{2}{c|}{Engine 15} & \multicolumn{2}{c}{All engines}  \\ 
  method &   RMSE & \multicolumn{1}{c|}{MAE} & RMSE & \multicolumn{1}{c|}{MAE}  & RMSE & \multicolumn{1}{c|}{MAE}  & RMSE & MAE     \\ 
     \hline
     FedAvg  & 8.8 & 7.2 & \textbf{12.8} & \textbf{9.2}  & \textbf{7.8} & 6.2 & \textbf{10.2} & \textbf{7.6}   \\ 
     \hline
     \multicolumn{9}{c}{Softmax aggregation policy} \\
     \hline
     Random-Softmax & 6.0 & 4.8 & 17.2 & 12.5 & 8.0 & \textbf{5.8} & 12.0 & 8.0   \\
    Full-Softmax  & \textbf{5.2} & \textbf{4.0} & 18.9 & 14.2 & 8.6 & 6.0 & 12.9 & 8.5  \\
     \hline
     \multicolumn{9}{c}{Best model aggregation policy} \\
     \hline
     Random-Best  & 6.8 & 5.8 & 13.5 & 10.4 & 8.7 & 7.1 & 10.4 & 7.9  \\
     Full-Best & 5.6 & 4.5 & 16.2 & 11.2 & 9.3 & 6.5 & 11.7 & 7.7 
    \end{tabular}
       \label{tab:comp_faulty}
\end{table*}

Figure \ref{fig:rmse_noise_all} shows the RMSE of the test engines for a noise multiplier $\alpha$ ranging from 0 (no noise) to 2 (very large noise). Table  \ref{tab:comp_faulty} and \ref{tab:comp_faulty_10} show the corresponding test results for a noise multiplier of $\alpha=0.1$ and $\alpha=1$ respectively. The FedAvg method performs well when the noise is relatively small, i.e., when $\alpha \leq 0.5$. For these noise values, the RMSE of all test engines is quite small. For $\alpha=0$ (no noise) and $\alpha=0.1$ (small noise), the RMSE is even smaller with the FedAvg method, than with all proposed aggregation methods. When the noise increases, however, i.e., from $\alpha=0.7$ onwards, the RMSE of the test engines increases as well (see Figure \ref{fig:rmse_noise_all}).

With the Softmax aggregation policies, the RMSE over all test engines is relatively high when $\alpha$ is small (see Figure \ref{fig:rmse_noise_softmax}). Table \ref{tab:comp_faulty} shows that this is mainly because for small $\alpha$ and the Softmax aggregation policies, the predictions for test engine 14 are inaccurate. When $\alpha$ increases, however, the RMSE over all test engines remains stable, and even decreases compared to the smaller values of $\alpha$. This is expected since, for large values of $\alpha$, it is easier to detect the clients with very noisy data, and  thus their contribution to the overall model decreases. For $\alpha=0.7$ onwards,  a lower RMSE is obtained over all test engines for both the Full-Softmax and the Random-Softmax method. Here, the RMSE over all test engines with the Full-Softmax method, is lower and more stable with the Random-Softmax method. This is as expected since the Random-Softmax method has an extra source of randomness, compared to the Full-Softmax method, because local models are randomly assigned to clients. This leads to less robust outcomes.

The outcomes with the best model aggregation policies  vary  with different level of noise. 
This is because for both Full-Best and the Random-Best method, the impact of randomness is larger, since the best model is selected as the new global model in each iteration. A small, random change in the training process, for instance slightly different initial weights, might already give another best model with the Full-Best and Random-Best methods.

Table \ref{tab:selected_models} shows how often the parameters (model weights) of each airline are selected as the parameters of the new global model with the Random-Best aggregation method. For a noise multiplier of $\alpha=0.1$ added to the data of engines 5 and 18, engine 5 is selected 3 times, while engine 18 is selected 17 times out of 50. In contrast, for a higher noise multiplier of $\alpha=1.0$ added to the data of engines 5 and 18, engine 5 is no longer selected, while engine 18 is only selected 8 times out of 50. This shows that the magnitude of the noise present in an engine's data impacts the frequency at which the engine is selected for the global model, which is in line with the results in Tables \ref{tab:comp_faulty} and \ref{tab:comp_faulty_10}. Notably, the engine that performs the best when trained in isolation (Engine $16$ in Table 5), is also one of the engines selected most frequently in both noise conditions. This is a desirable outcome, as training a model in islotation with Engine 16’s data provided good test results, as shown in Table \ref{tab:predictions_cen_vs_iso}. Overall, Table \ref{tab:selected_models} shows the variety in the contributions of all models trained on different engine datasets, in spite of their dissimilarity to the validation dataset data distributions, promoting the inclusion of features from various clients.

\begin{table*}[!ht]
    \centering
        \caption{Comparison of the RMSE and MAE (in flights) of the RUL prognostics for all three test engines in N-CMAPSS, for all considered  aggregation methods. Here, additional noise with a standard deviation multiplier of $\alpha=1$ is added to the data of engine 5 and 18. The best results are denoted in bold. }
    \begin{tabular}{c|cccccc|cc}
   Aggregation & \multicolumn{2}{c|}{Engine 11} & \multicolumn{2}{c|}{Engine 14} & \multicolumn{2}{c|}{Engine 15} & \multicolumn{2}{c}{All engines}  \\ 
  method &   RMSE & \multicolumn{1}{c|}{MAE} & RMSE & \multicolumn{1}{c|}{MAE}  & RMSE & \multicolumn{1}{c|}{MAE}  & RMSE & MAE     \\ 
     \hline
     FedAvg  & 13.8 & 11.8 & \textbf{11.9} & 9.7 & 11.5 & 9.5 & 12.3 & 10.2 \\ 
     \hline
     \multicolumn{9}{c}{Softmax aggregation policy} \\
     \hline
     Random-Softmax & 7.6 & 6.3 & 13.9 & 9.7 & 8.0 & 6.3 & 10.5 & 7.6 \\
       Full-Softmax   & 5.9 & 5.0 & 13.7 & \textbf{9.5} & 8.0 & \textbf{5.9} & 10.1 & 7.0 \\
     \hline
     \multicolumn{9}{c}{Best model aggregation policy} \\
     \hline
     Random-Best  & \textbf{5.6} & \textbf{4.4} & 19.1 & 14.2 & 10.1 & 7.1 & 13.4 & 9.0 \\
     Full-Best &  6.1 & 5.1 & 12.6 & 9.1 & \textbf{7.7} & \textbf{5.9} & \textbf{9.5} & \textbf{6.9} 
    \end{tabular}

    \label{tab:comp_faulty_10}
\end{table*}

\begin{table*}[!ht]
    \centering
    \caption{\textcolor{black}{An overview of how often the parameters (model weights) of each airline are selected as the parameters of the new global model, and in which epochs these airlines are selected, for the Random-Best aggregation method with $\alpha=0.1$ (small noise) and $\alpha=1.0$ (large noise). }  }
    \begin{tabular}{cc|c|c}
            &                     & $\#$ Times & \\
        Noise Multiplier &  Airline (Engine) & Selected & Selected in epochs \\
        \hline
        \multirow{6}{2cm}{$\alpha=0.1$} & Airline A (Engine 2)& 7 &2, 3, 19, 20, 27, 40, 49  \\
        & Airline B (Engine 5)& 3 & 8, 24, 50  \\
        & Airline C (Engine 10)& 10 & 4, 11, 12, 17, 34, 35, 38, 43, 46, 47,     \\
        & Airline D (Engine 16)& 12 & 7, 10, 21, 25, 26, 30, 31, 32, 36, 41, 45, 48        \\
        & Airline E (Engine 18)& 17 &1, 5, 6, 9, 13, 14, 15, 16, 18, 22, 23, 28, 29, 33, 37, 39, 42    \\
        & Airline F (Engine 20)& 1 & 44 \\
        \hline
        \multirow{6}{2cm}{$\alpha=1.0$} & Airline A (Engine 2)& 3 & 8, 13, 35  \\
        & Airline B (Engine 5)& 0 & - \\
        & Airline C (Engine 10)& 15 & 1, 3, 4, 10, 12, 14, 16, 22, 23, 25, 31, 32, 37, 39, 44    \\
        & Airline D (Engine 16)& 17 & 9, 11, 15, 19, 26, 27, 28, 30, 33, 34, 36, 38, 41, 42, 45, 47, 48 \\
        & Airline E (Engine 18)& 8 & 6, 17, 21, 24, 29, 46, 49, 50 \\
        & Airline F (Engine 20)& 7 & 2, 5, 7, 18, 20, 40, 43 
    \end{tabular}
    \label{tab:selected_models}
\end{table*}

\subsection{\textcolor{black}{Limitations of this work}}

\paragraph{\textcolor{black}{Privacy vs Robustness in the aggregation methods}}
\textcolor{black}{The proposed aggregation methods enhance  robustness to excessive noise in the data of some clients. For these  aggregation methods, the server shares the weights of the local model of an airline with the other airlines. 
However, information about the training data of each airline is stored in its local weights  \cite{alebouyeh2024benchmarking}. An airline could therefore perform inversion attacks and property inference attacks to infer information about, and characteristics of, the training data of another airline from the model weights \cite{alebouyeh2024benchmarking, zhu2019deep, geiping2020inverting}. There is thus a trade-off between the robustness of the approach and the privacy protection of the airlines.}

{\color{black}
\paragraph{The impact the number of clients with noisy data on the performance of the proposed aggregation methods.} 
In the context of distributed learning, it is expected that noise may appear in some of the client's (airline's) datasets. Nevertheless, given that the methods presented in this paper rely on the statistical median's robustness to outliers, as soon as more than half the clients contain noise in their data, the median can no longer be used as a suitable selector. Therefore, these methods are limited to contexts where no more than half the clients will present noise in their datasets. As a result, different approaches should be considered for such cases.
}

\paragraph{\textcolor{black}{Approval from the regulatory authorities}}
\textcolor{black}{Prognostic models for aircraft have to be verified and approved by regulatory authorities such as the  European Union Aviation Safety Agency (EASA, \cite{EASA}) in Europe, or the Federal Aviation Administration (FAA) in the USA.  
For this, it might be necessary that airlines share aggregated information about their data.
In general, however, a single airline may not have enough data to get a prognostic model approved by the regulatory authorities. The methodology proposed in this paper is a first attempt to  develop a collaborative framework towards getting approval from the regulatory authorities.}

\paragraph{\textcolor{black}{Stragglers}}
\textcolor{black}{Another limitation of the proposed framework is the lack of support to mitigate stragglers in a heterogeneous setup~\cite{9778210}, where the speed of the model training is determined by the slowest device. This might hinder the speed of training of the RUL prognostic model. Nevertheless, it is expected that airlines are interested in investing in appropriate hardware and software capabilities to support a  scalable implementation of the proposed FL framework.}




\section{Conclusion}\label{sec:Conclusion}

In this paper, a feasible federated learning framework is developed that enables airlines to collaboratively train a common machine learning model for RUL prognostics, without the need to share their data. Validating the common machine learning model for RUL prognostics is crucial for airlines to be able to certify and implement such model in practice. However, to address the reluctancy to collaborate due to data sharing and potential noisy data from peers, a decentralised validation procedure and four aggregation methods are proposed within the FL framework.

The proposed FL framework is applied for the development of RUL prognostic models for aircraft engines \cite{arias2021aircraft}. Six airlines are considered. Using the proposed FL framework, accurate RUL prognostics are obtained, with a RMSE of 9.9 flights. The performance of the FL framework is compared with the case where airlines do not cooperate, i.e.,  each airline is independently training its own RUL prognostic model. In this case, the mean RMSE (over all six airlines) is 15.8 flights, showing a significant decrease in the accuracy of the RUL prognostics. The FL approach provides more accurate RUL prognostics for five out of the six considered airlines. Overall, the results show that the proposed FL framework gives accurate RUL prognostics, while the privacy of their data is preserved. Lastly, four novel aggregation methods are proposed for the global model parameters. When all airlines have no or little noise in the data, the proposed methods and the standard FedAvg method have a comparable performance. However, when some airlines have  noisy condition monitoring data, the proposed aggregation methods provide more accurate RUL prognostics. 

Overall, this paper  investigates the robustness property of FL, allowing the construction of effective global models, even in the presence of noise. However, as pointed out in \cite{alebouyeh2024benchmarking, zhu2019deep, geiping2020inverting}, FL procedures require special attention to protect a client's privacy. As such, our future work considers the development of such procedures {\color{black} and explore privacy-preserving collaborative learning mechanisms}, while at the same time remaining robust to noisy data. Such approaches will also consider authentication as an additional measure to improve the robustness of aggregation methods. Nevertheless, this paper has shown how multiple airlines would benefit in collaborating for the development of a global RUL prognostics model. At the same time, these methods ensure a minimal impact on the resulting model from clients with noisy data.

\bibliography{references}
\renewcommand{\thesection}{\Alph{section}}
\setcounter{section}{0}
{\color{black}
\section{Appendix: Artifact Description}\label{sec:appendix}
We briefly present in this section the reproducibility artifact of our
proposed FL framework.
\subsection{Release}
The complete prototype of the proposed FL framework and its source code is released open-source~\cite{rul-fl} in an \emph{online GitHub} repository at ~\url{https://github.com/EC-labs/rul-fl}.

\subsection{Hardware Dependencies}
We implemented and tested the proposed FL framework on an \verb|i7-11850H Intel| server @ \SI{2.50}{\giga\hertz} with $8$ CPU cores, $2$$\times$\SI{16}{\giga\byte} \verb|DDR4| RAM, and an encryption enabled \SI{1}{\tera\byte} \verb|SSD| disk. The software, however, is portable and has no hardware dependencies.
\subsection{Software Dependencies}
We tested the proposed FL framework on an Ubuntu 22.04 LTS, Jammy Jellyfish (\verb|x86_64|) OS with \verb|6.8.12| Linux kernel, however it runs on any Linux-based environment. The implemented framework has no other software dependencies.
\subsection{Dataset}
The complete dataset~\cite{turbofan-dataset} used for illustrating the usefulness of the proposed FL framework is available at ~\url{https://phm-datasets.s3.amazonaws.com/NASA/17.+Turbofan+Engine+Degradation+Simulation+Data+Set+2.zip}. The data preprocessing details are presented in Section~\ref{sec:data}.
\subsection{Evaluation Data}
The complete data used for analysis in the paper is available at \emph{GitHub} repository~\cite{rul-fl} in the compressed format, and can be found in the \verb|results| directory as:
\begin{lstlisting}
    ./results/evaluation=2024-03-14.zip
\end{lstlisting}

Reproducing the analysis data in the \verb|results| directory requires executing \verb|run.sh| script in the \emph{GitHub} repository~\cite{rul-fl}. Currently, as part of FL configuration, the \verb|run.sh| script runs $5$ clients and $1$ server for each federated learning algorithm (i.e., fedavg, full-softmax, full-best, random-softmax, random-best) presented in the paper. The \verb|run.sh| script also runs experiments for the unrestricted access centralized, and non-collaborative isolated learning scenarios.

Additionally, the \verb|run.sh| script also runs experiments for multiple noise configurations, starting at $\alpha=0.1$ to $\alpha=2.0$. The analysis data related to complete experimentation can be found in the \verb|results/| directory.
}

\end{document}